\documentclass{article}

\usepackage{microtype}
\usepackage{graphicx}
\usepackage{subfigure}
\usepackage{booktabs}
\usepackage{hyperref}

\usepackage[accepted]{icml2024}

\usepackage{amsmath}
\usepackage{amssymb}
\usepackage{mathtools}
\usepackage{amsthm}
\usepackage{mwe}
\usepackage{pifont}
\usepackage{wrapfig}
\usepackage{amssymb}
\usepackage{colortbl}
\usepackage{multirow}
\usepackage{multicol}
\usepackage{dsfont}

\usepackage{makecell}
\usepackage{bbm}

\newcommand{\eg}{\textit{e}.\textit{g}.}
\newcommand{\ie}{\textit{i}.\textit{e}.}

\newcommand*{\QEDA}{\null\nobreak\hfill\ensuremath{\blacksquare}}

\definecolor{jleecolor}{rgb}{0.65,0.16,0.65}

\usepackage[capitalize,noabbrev]{cleveref}

\theoremstyle{plain}

\theoremstyle{definition}

\theoremstyle{remark}

\usepackage[textsize=tiny]{todonotes}

\icmltitlerunning{3D Geometric Shape Assembly via Efficient Point Cloud Matching}

\begin{document}

\twocolumn[
\icmltitle{3D Geometric Shape Assembly via Efficient Point Cloud Matching}

\icmlsetsymbol{equal}{*}

\begin{icmlauthorlist}
\icmlauthor{Nahyuk Lee}{equal,postech_cse}
\icmlauthor{Juhong Min}{equal,postech_ai}
\icmlauthor{Junha Lee}{postech_cse}
\icmlauthor{Seungwook Kim}{postech_ai}
\icmlauthor{Kanghee Lee}{snu_cse}
\icmlauthor{Jaesik Park}{snu_cse,snu_ai}
\icmlauthor{Minsu Cho}{postech_cse,postech_ai}
\end{icmlauthorlist}

\icmlaffiliation{postech_cse}{Department of Computer Science and Engineering, POSTECH, Pohang, Korea}
\icmlaffiliation{postech_ai}{Graduate School of Artificial Intelligence, POSTECH, Pohang, Korea}
\icmlaffiliation{snu_cse}{Department of Computer Science and Engineering, Seoul National University, Seoul, Korea}
\icmlaffiliation{snu_ai}{Interdisciplinary Program in Artificial Intelligence, Seoul National University, Seoul, Korea}

\icmlcorrespondingauthor{Minsu Cho}{mscho@postech.ac.kr}

\icmlkeywords{Geometric shape assembly, High-dimensional feature transform, Correlation aggregation, Proxy Match Transform}

\vskip 0.3in
]



\printAffiliationsAndNotice{\icmlEqualContribution} 

\begin{abstract}
Learning to assemble geometric shapes into a larger target structure is a pivotal task in various practical applications.
In this work, we tackle this problem by establishing local correspondences between point clouds of part shapes in both coarse- and fine-levels. 
To this end, we introduce Proxy Match Transform (PMT), an approximate high-order feature transform layer that enables reliable matching between mating surfaces of parts while incurring low costs in memory and computation. 
Building upon PMT, we introduce a new framework, dubbed Proxy Match TransformeR (PMTR), for the geometric assembly task.
We evaluate the proposed PMTR on the large-scale 3D geometric shape assembly benchmark dataset of Breaking Bad and demonstrate its superior performance and efficiency compared to state-of-the-art methods. Project page: \url{https://nahyuklee.github.io/pmtr}.
\end{abstract}


\section{Introduction}
Shape assembly aims to determine the precise placement of each constituent part and construct a larger target shape as a whole.
This task holds paramount significance, especially in the context of various applications encompassing robotics~\cite{wang2019stable,zakka2020form2fit,zeng2021transporter}, manufacturing~\cite{tian2022assemble}, computer graphics~\cite{Li_siga12}, and computer-aided design~\cite{chen2015dapper,jacobson2017generalized}. 
Despite its pivotal role in industrial productivity and the plethora of applications, the field of shape assembly remains relatively underexplored in the literature due to the intricate challenge it presents:
demands a comprehensive understanding of geometric structures and analyses of pairwise relationships between local surfaces of input parts to establish accurate assembly.

There have been several recent attempts~\cite{schor2019componet, li2020global, wu2020lstm, li2020learning, huang2020dgl, narayan2022rgl, chen2022neural, wu2023leveraging} to address the task of shape assembly, but these methods fall short of achieving accurate assembly. 
They typically represent each part as a global embedding and perform regression to predict a placement for each part. 
The global encoding strategy for each part, while simplifying the process, greatly limits local information by collapsing spatial resolutions, which is necessary to localize the mating surface.
Indeed, accurate shape assembly requires a detailed analysis of both fine- and coarse-level spatial information of the parts in recognizing mating surfaces and establishing correspondences between the surfaces.
Therefore, a promising approach would be to retain the spatially rich part representations during the encoding phase and analyze pairwise local correspondence relationships between them for reliable localization and matching of mating surfaces.

In the realm of correspondence analysis within image matching, prior methods~\cite{rocco2018neighbourhood,min2021chm,swkim2022tfmatcher,min2021hypercorrelation,rocco2020sparsencnet} typically utilize a high-order feature transform, \ie, high-dimensional convolution or attention, to achieve the objectives of localizing relevant instances and establishing correspondences between them.
The high-order feature transforms, which assess structural patterns of correlations in high-dimensional spaces, have been empirically validated for their efficacy in identifying accurate visual matches.
However, the quadratic complexity with respect to input spatial resolution still remains as a significant drawback, {\em limiting their application to only low-resolution (coarse-grained) inputs}.
Such a limitation becomes particularly problematic in the context of geometric assembly since meticulous alignment between parts {\em requires analyzing high-resolution (fine-grained)} to precisely identify `geometric compatibility' between mating surfaces to match.

\begin{figure*}[t]
\begin{center}
\includegraphics[width=0.9\textwidth]{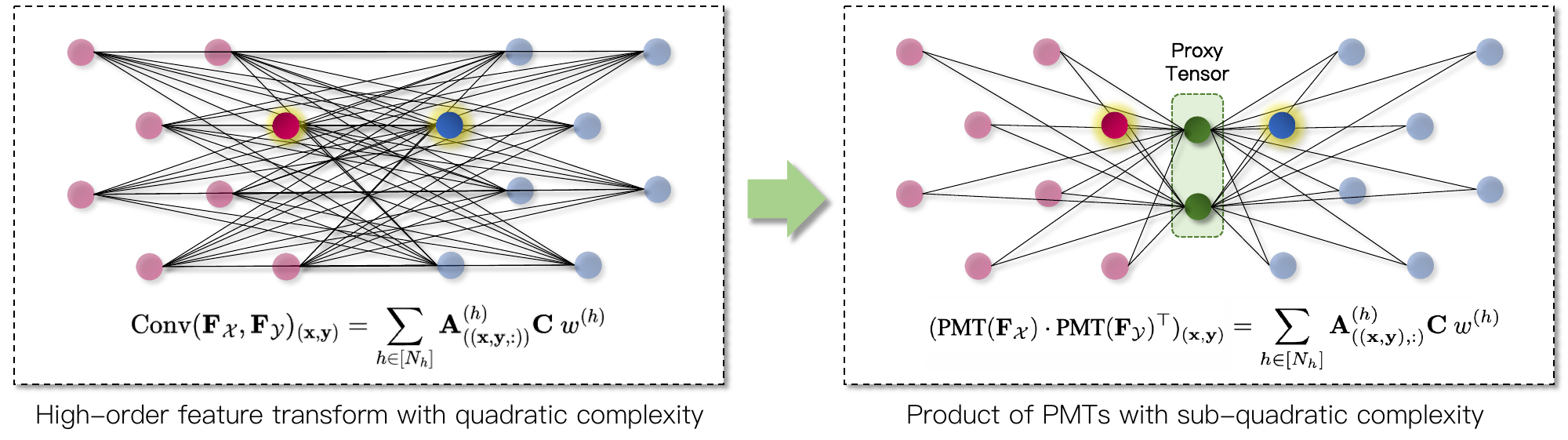}
\vspace{-3mm}
\caption{Given a correlation score at position $(\mathbf{x}, \mathbf{y})$ (the edge between highlighted nodes) and its neighboring scores (all other edges), vanilla high-order feature transform (shown on the left) leads to quadratic complexity due to its demand for memory-intensive pairwise correlation scores.
The product of two PMTs (shown on the right) effectively approximates this high-order transform only with sub-quadratic complexity by avoiding direct construction of correlation scores, instead exchanging information through a low-dimensional proxy tensor.
The red/blue nodes and black edges represent the source/target features and the correlation scores between them, respectively.}
\label{fig:teaser}
\end{center}
\vskip -0.2in
\end{figure*}

In this paper, we address this issue by introducing a new form of low-complexity high-order feature transform layer, dubbed \textit{Proxy Match Transform (PMT)}, to tackle the challenges of geometric shape assembly.
The layer is designed to align analogous local embeddings in feature space, \eg, points on mating surfaces, with sub-quadratic complexity, thus offering a low-complexity yet high-order approach as illustrated in Fig.~\ref{fig:teaser}.
We theoretically prove that the proposed PMT layer can effectively approximate the conventional high-order feature transforms~\cite{rocco2018neighbourhood,choy2020deep,min2021chm} under particular conditions.
To demonstrate its efficacy, we incorporate the PMT layer into a coarse-to-fine matching framework, Proxy Match TransformeR (PMTR), that uses PMTs for both coarse- and fine-level matching steps to establish correspondences on mating surfaces.
We compare our results with the recent state of the arts and provide a thorough performance analysis on the standard geometric shape assembly benchmark of Breaking Bad~\cite{sellan2022breakingbad}.
The experiments demonstrate that our method significantly outperforms existing approaches while being computationally efficient compared to the baselines.

Our main contributions can be summarized as follows:
\begin{itemize}
\vspace{-2.0mm}
\setlength\itemsep{+0.2em}
    \item We introduce Proxy Match Transform (PMT), a low-complexity high-order feature transform layer that effectively refines the matching of the feature pair.
    \item Our theoretical analysis shows that PMT effectively approximates high-order feature transform while incurring sub-quadratic memory and time complexity.
    \item Performance improvements in geometric shape assembly over the state-of-the-art baselines demonstrate the effectiveness and efficiency of our approach.
\end{itemize}

\section{Related Work}

\smallbreak
\noindent \textbf{3D shape assembly \& registration.}
Previous research in generative models for 3D objects has primarily focused on building objects through the combination of basic 3D primitives.
A prevalent approach trains specialized models tailored to individual object classes, enabling the assembly of objects from volumetric primitives such as cuboids \citep{tulsiani2017learning}.
In contrast, \citet{khan2019unsupervised} proposes a unified model that can generate cuboid primitives in various classes.
Additionally, variational autoencoders (VAEs) have been employed to model objects as compositions of cuboids, offering robust abstractions that distill local geometric details and elucidate object correspondences \citep{kingma2014auto, jones2020shapeassembly}.

Parallel to these developments, research in the part assembly has aimed to construct complete objects from predefined semantic parts.
The method of \citet{li2020learning} predicts translations and rotations for part point clouds to assemble a target object from an image reference.
Extending this, \citet{narayan2022rgl, huang2020dgl} have conceptualized part assembly as a graph learning challenge, utilizing iterative message passing techniques to integrate parts into cohesive objects.
These approaches heavily rely on the PartNet dataset \citep{mo2019partnet} to ensure semantic correspondence between the assembled parts and the target models, demonstrating that while geometric shapes are foundational, semantic cues can significantly guide and streamline the assembly process.
Our research diverges from these methods by focusing on the assembly of parts without predefined semantics.
A closely related methodology is that of \citet{chen2022neural}, which also tackles the problem of 3D shape assembly by integrating implicit shape reconstruction, providing a relevant benchmark.

Additionally, the concept of 3D shape assembly overlaps with the domain of 3D registration, especially in scenarios characterized by low overlap between a pair of point clouds.
Techniques such as those proposed by \citet{huang2021predator} and \citet{yu2021cofinet} leverage self-attention and cross-attention mechanisms within and across point cloud features to transform 3D features, facilitating enhanced matching accuracy.
\citet{qin2022geotransformer} further advances this by mapping transformation-invariant data to the positional embeddings of transformer layers, optimizing the matching process in low-overlap conditions.
Despite their efficacy, the practical application of these methods in fine-grained matching scenarios is often constrained by the {\em quadratic complexity} associated with their matching layers, highlighting a critical area for improvement in computational efficiency and scalability.
Our work addresses these challenges by proposing a novel approach that optimizes the computational demands of feature matching while maintaining high robustness.

\smallbreak 
\noindent \textbf{High-order feature transform for matching.}
High-order feature transforms are essential in (both image and point cloud) matching tasks, helping to establish consensus among correspondences within a high-dimensional space.
Initially introduced by \citet{rocco2018neighbourhood}, the concept of a learning-based neighborhood consensus supports the identification of accurate matches by leveraging neighboring ambiguous matches between 2D images.
This approach has also been adapted for 3D registration tasks, notably by \citet{choy2020deep}, who utilized a 6D sparse convolutional layer to filter out outlier correspondences.
Given the high computational complexity associated with high-order feature transforms, several studies have proposed methods to reduce this burden.
Techniques such as decomposing high-dimensional convolutional kernels \citep{min2021hypercorrelation} and sparsifying the correlation map with top-$k$ scores \citep{rocco2020sparsencnet} have been effective.
Further, \citet{shi2023clustergnn} enhanced matching efficiency by creating a sparse correlation matrix through the grouping of input tokens, significantly reducing the number of tokens involved.
More recent advances have integrated the self-attention mechanism to utilize global feature consensus effectively, although these methodologies, proposed by \citet{cho2021cats} and \citet{swkim2022tfmatcher}, come at a higher computational cost.

Our work introduces the Proxy Match Transform (PMT), which simplifies existing high-order feature transforms to significantly reduce computational demands.
We apply PMT in a coarse-to-fine approach, identifying reliable correspondences between the mating surfaces of input parts and subsequently refining them for precise assembly.
There have been several approaches relevant to ours such as leveraging local geometric cues for assembly by~\citet{lu2023jigsaw}, the linear approximations in convolutional networks by \citet{denton2014exploiting}, sparse attention mechanisms by \citet{zaheer2021big}, low-rank approximations of self-attention by \citet{chen2020compressed}, and Gaussian kernel approximations by \citet{chen2021skyformer}.
However, unlike these methods, which primarily improve processing within a {\em single feature}, PMT uniquely addresses the challenge of efficient matching between {\em two distinct features}, improving both computational efficiency and feature correspondence analysis, which are essential for diverse applications including geometric shape assembly.


\section{Proposed Approach}

In the task of geometric shape assembly, analyzing geometric compatibility between fractured shapes is of utmost importance;
the geometric properties of the {\em mating surfaces} should exhibit consistency, where vertices, edges, and surfaces seamlessly fit together to form a coherent structure.
To achieve reliable localization of mating surfaces between shapes, a model needs to analyze the compatibility of all possible feature correspondences and accurately identify spatially consistent matches. 
In the field of visual matching and its applications~\citep{rocco2018neighbourhood,choy2020deep,min2021chm,cho2021cats,min2021hypercorrelation}, a trending approach for assessing match reliability is the utilization of \textbf{\textit{high-order feature transform}}, \eg, convolution or self-attention.
This technique effectively assesses patterns within neighborhood matches in a differentiable manner.
Building upon these principles, we will now explore the theoretical formulation of high-order transform, with a specific emphasis on its application for enhancing pairwise feature correlation.

\smallbreak
\noindent \textbf{Preliminary.} High-order convolution~\citep{rocco2018neighbourhood,choy2020deep,min2021chm} generalizes the standard convolution by taking as input more functions, feature maps, or sets.
In the context of our problem, we consider two point clouds $\mathcal{X} = \{ \mathbf{x}_{i} \in \mathbb{R}^{3} \}_{i=1}^{N}$ and $\mathcal{Y} = \{ \mathbf{y}_{i} \in \mathbb{R}^{3} \}_{i=1}^{M}$, and focus on the 2nd-order convolution with two sets of features $\mathcal{F}_\mathcal{X}$ and $\mathcal{F}_\mathcal{Y}$, associated with the two point clouds, respectively.
For ease of notation, we represent these features in matrix form, \ie,  $\mathbf{F}_\mathcal{X} \in \mathbb{R}^{|\mathcal{X}| \times D_{\text{emb}}}$, where $D_{\text{emb}}$ is the feature embedding dimension, and indexes each feature embedding using its associated point $\mathbf{x} \in \mathcal{X}$ such that $(\mathbf{F}_{\mathcal{X}})_\mathbf{x} \in \mathbb{R}^{D_{\text{emb}}}$, and same goes for $\mathcal{F}_\mathcal{Y}$.
We also express the feature correlation of two points from each point cloud, $\mathbf{x}$ and $\mathbf{y}$,  
as $\mathbf{C}_{(\mathbf{x}, \mathbf{y})} \coloneqq (\mathbf{F}_{\mathcal{X}})_{\mathbf{x}} \cdot (\mathbf{F}_{\mathcal{Y}})_{\mathbf{y}}^{\top} \in \mathbb{R}$. 
The 2nd-order convolution on $(\mathbf{F}_\mathcal{X}, \mathbf{F}_\mathcal{Y})$ with kernel $K$ is then defined as:
\begin{multline}
    \text{Conv}(\mathbf{F}_\mathcal{X}, \mathbf{F}_\mathcal{Y})_{(\mathbf{x}, \mathbf{y})} \coloneqq  \\ \sum_{(\mathbf{n}, \mathbf{m}) \in \mathcal{N}(\mathbf{x}) \times \mathcal{N}(\mathbf{y})} \mathbf{C}_{(\mathbf{n}, \mathbf{m})} {K}([\mathbf{n} - \mathbf{x}, \mathbf{m}-\mathbf{y}]),
    \label{eq:hdc}
\end{multline}
where $\mathcal{N}(\cdot)$ represents a set of neighbor points and $K: \mathbb{R}^{6} \xrightarrow{} \mathbb{R}$ is a convolutional kernel, represented as a mapping function that takes a displacement vector onto learnable weight scalar.

Building upon insights from the work of~\citet{cordonnier2020relationship}, we consider Lemma 1 which states that the conv layer in Eq.~\ref{eq:hdc} can be re-formulated as a form of multi-head self-attention under sufficient conditions:
\smallbreak
\noindent \textbf{Lemma 1.} {\em Consider a bijective mapping of natural numbers, i.e., heads, onto 6-dimensional local displacements: $t(h): [{N_h}] \rightarrow \Delta(\mathbf{x}, \mathbf{y})$. Let ${\mathbf{A}}^{(h)} \in \mathbb{R}^{|\mathcal{X}||\mathcal{Y}| \times |\mathcal{X}||\mathcal{Y}|}$ be an attention matrix that holds the following:}
\begin{align}
    {\mathbf{A}}^{(h)}_{(\mathbf{x}, \mathbf{y}), (\mathbf{n}, \mathbf{m})} = \begin{cases} \mbox{1,} & \mbox{if $t(h) = (\mathbf{n}, \mathbf{m}) - (\mathbf{x}, \mathbf{y})$} \\ \mbox{0,} & \mbox{otherwise.} \end{cases} \label{eq:local_constraint_2s}
\end{align}
{\em Then, for any high-dimensional convolution with a kernel $K: \mathbb{R}^{6} \xrightarrow{} \mathbb{R}$, there exists $\{w^{(h)} \in \mathbb{R}\}_{h \in [N_h]}$ such that following equality holds:}
\begin{align}
    \text{Conv}(\mathbf{F}_{\mathcal{X}}, \mathbf{F}_{\mathcal{Y}})_{(\mathbf{x}, \mathbf{y})} = \sum_{h \in [{N_h}]} {\mathbf{A}}^{(h)}_{((\mathbf{x}, \mathbf{y}), :)} \mathbf{C} \ {w}^{(h)}.
    \label{eq:hdc_att}
\end{align}

{\em Proof.} We refer to the Appendix~\ref{sec:theorem} for the complete proof.

As illustrated in the left of Fig.~\ref{fig:teaser}, the 2nd-order convolution (Eq.~\ref{eq:hdc} and~\ref{eq:hdc_att}) is designed to disambiguate spatially consistent correspondences and update their correlation values by analyzing local correlation patterns around each point pair $(\mathbf{x}, \mathbf{y}) \in \mathcal{X} \times \mathcal{Y}$.
Despite its good empirical performance in literature~\citep{rocco2018neighbourhood,choy2020deep,min2021chm,min2021hypercorrelation}, its critical limitation lies in the quadratic complexity of correlation computation, \ie, $\mathcal{O}(|\mathcal{X}|\cdot|\mathcal{Y}|)$, with respect to the input resolution, imposing significant computational burdens during both the training and inference phases.
This restricts its practical applications with large spatial resolution inputs, such as the geometric shape assembly task that demands high-resolution, \ie, geometric-level, input processing for geometric compatibility analysis to ensure precise correspondence alignments.

\subsection{Proxy Match Transform: an efficient high-order feature transform with sub-quadratic complexity}
\label{sec:proxymatch}
To overcome the limitation, we introduce an efficient feature matching layer, dubbed \textit{Proxy Match Transform}, which approximates high-order convolution with sub-quadratic complexity.
Given a pair of features $(\mathbf{F}_\mathcal{X}, \mathbf{F}_\mathcal{Y})$ as inputs, PMT layers with $N_h$ heads\footnote{Similar to multi-head self-attention~\citep{vaswani2017attention}, each head performs distinct attentions and feature transform, allowing the layer to attend different aspects of inputs.} are defined as follows:
\begin{align}
    \text{PMT}(\mathbf{F}_{\mathcal{X}}) &\coloneqq \sum_{h \in [{N_h}]} \mathbf{A}_{\mathcal{X}}^{(h)} \mathbf{F}_{\mathcal{X}} \mathbf{P}^{(h)\top} {w}_{\mathcal{X}}^{(h)}, \\ 
    \text{PMT}(\mathbf{F}_{\mathcal{Y}}) &\coloneqq \sum_{h \in [{N_h}]} \mathbf{A}_{\mathcal{Y}}^{(h)} \mathbf{F}_{\mathcal{Y}} \mathbf{P}^{(h)\top} {w}_{\mathcal{Y}}^{(h)},
    \label{eq:proxymatch}
\end{align}
where $w_{\mathcal{X}}^{(h)} \in \mathbb{R}$ is a learnable weight scalar, $\mathbf{A}_{\mathcal{X}}^{(h)} \in \mathbb{R}^{|\mathcal{X}| \times |\mathcal{X}|}$ is local attention matrix\footnote{To avoid the quadratic complexity of $|\mathcal{X}| \times |\mathcal{X}|$ in the attention matrices, we adopt an implementation strategy similar to that described in~\citet{thomas2019kpconv}. We refer to Sec.~\ref{sec:impl} for details.}, and $\mathbf{P}^{(h)} \in \mathbb{R}^{D_{\text{proxy}} \times D_{\text{emb}}}$ is \textbf{proxy tensor} that satisfies the following:
\begin{align}
    \mathbf{P}^{(i)\top} \mathbf{P}^{(j)} = \begin{cases} \mbox{$\mathbf{I}_{D_{\text{emb}}}$,} & \mbox{if $i=j$} \\ \mbox{$\mathbf{0}$,} & \mbox{otherwise.} \end{cases} \label{eq:local_constraint}
\end{align}
where $D_{\text{proxy}}$ refers to the spatial resolution of the proxy tensor satifying $D_{\text{proxy}} \ll |\mathcal{X}|, |\mathcal{Y}|$.
The constraint ensures orthogonality between different proxy tensors.
The rationale behind this design is discussed in Sec.~\ref{sec:constraint}.

At each head, the layer initially constructs a correlation between the input feature $\mathbf{F}_{\mathcal{X}}$ and the proxy tensor $\mathbf{P}^{(h)}$ such that $\mathbf{C}_{\mathcal{X}}^{(h)} \coloneqq \mathbf{F}_{\mathcal{X}}\mathbf{P}^{(h)\top}$ in much smaller size of $|\mathcal{X}| \times D_{\text{proxy}}$, compared to the pairwise feature correlation $\mathbf{C} = \mathbf{F}_{\mathcal{X}}{\mathbf{F}_{\mathcal{Y}}}^{\top} \in \mathbb{R}^{|\mathcal{X}| \times |\mathcal{Y}|}$ as defined in Eq.~\ref{eq:hdc}.
After applying learnable weight ${w}_{\mathcal{X}}^{(h)}$, the output at position $(\mathbf{n}, \mathbf{m}) \in |\mathcal{X}| \times D_{\text{proxy}}$ is computed through a weighted-sum of its neighborhood matches lying on the spatial dimension of feature map $\mathbf{F}_{\mathcal{X}}$, \eg, $\{ (\mathbf{n}', \mathbf{m}) \}_{\mathbf{n}' \in \mathcal{N}(\mathbf{n})}$ where $|\mathcal{N}(\mathbf{n})| = \epsilon \ll |\mathcal{X}|$.
To formally put, the Proxy Match Transform output at head $h$ given input $\mathbf{F}_{\mathcal{X}}$ at position $(\mathbf{n}, \mathbf{m})$ is defined as
\begin{multline}
    {\text{PMT}(\mathbf{F}_{\mathcal{X}})}^{(h)}_{(\mathbf{n}, \mathbf{m})} = 
    {\mathbf{A}_{\mathcal{X}}^{(h)}}_{(\mathbf{n}, :)}\mathbf{F}_\mathcal{X}{\mathbf{P}^{(h)\top}}_{(:,\mathbf{m})} {w}_{\mathcal{X}}^{(h)}  \\
    = {\mathbf{A}_{\mathcal{X}}^{(h)}}_{(\mathbf{n}, :)} {\mathbf{C}_{\mathcal{X}}^{(h)}}_{(:, \mathbf{m})} {w}_{\mathcal{X}}^{(h)}  \\ 
    = \sum_{\mathbf{n}' \in \mathcal{N}(\mathbf{n})} {\mathbf{A}_{\mathcal{X}}^{(h)}}_{(\mathbf{n}, \mathbf{n}')} {\mathbf{C}_{\mathcal{X}}^{(h)}}_{(\mathbf{n}', \mathbf{m})} {w}_{\mathcal{X}}^{(h)}.
\end{multline}
$\text{PMT}(\mathbf{F}_{\mathcal{Y}})^{(h)}$ is similarly defined with a different set of parameters of $\mathbf{A}_{\mathcal{Y}}^{(h)}$ and ${w}_{\mathcal{Y}}^{(h)}$.

It is important to note that the PMT layers perform two \textbf{{\em independent}} transforms for \textbf{{\em feature matching}}, one for $\mathbf{F}_{\mathcal{X}}$ and the other for $\mathbf{F}_{\mathcal{Y}}$.
Despite the independence, matching between the feature pair is effectively facilitated by a shared proxy tensor $\mathbf{P}$. 
This proxy tensor allows for the exchange of information between the features, eliminating the need to construct and convolve memory-intensive pairwise feature correlations, which often contain sparse and limited informative match scores.
We demonstrate how the PMT effectively approximates existing high-order convolution in Sec.~\ref{sec:constraint} and empirically prove the efficacy of the use of proxy tensor and different parameter sets in geometric shape assembly in Sec.~\ref{sec:ablation}.

\begin{figure*}[!t]
\begin{center}
\includegraphics[width=\textwidth]{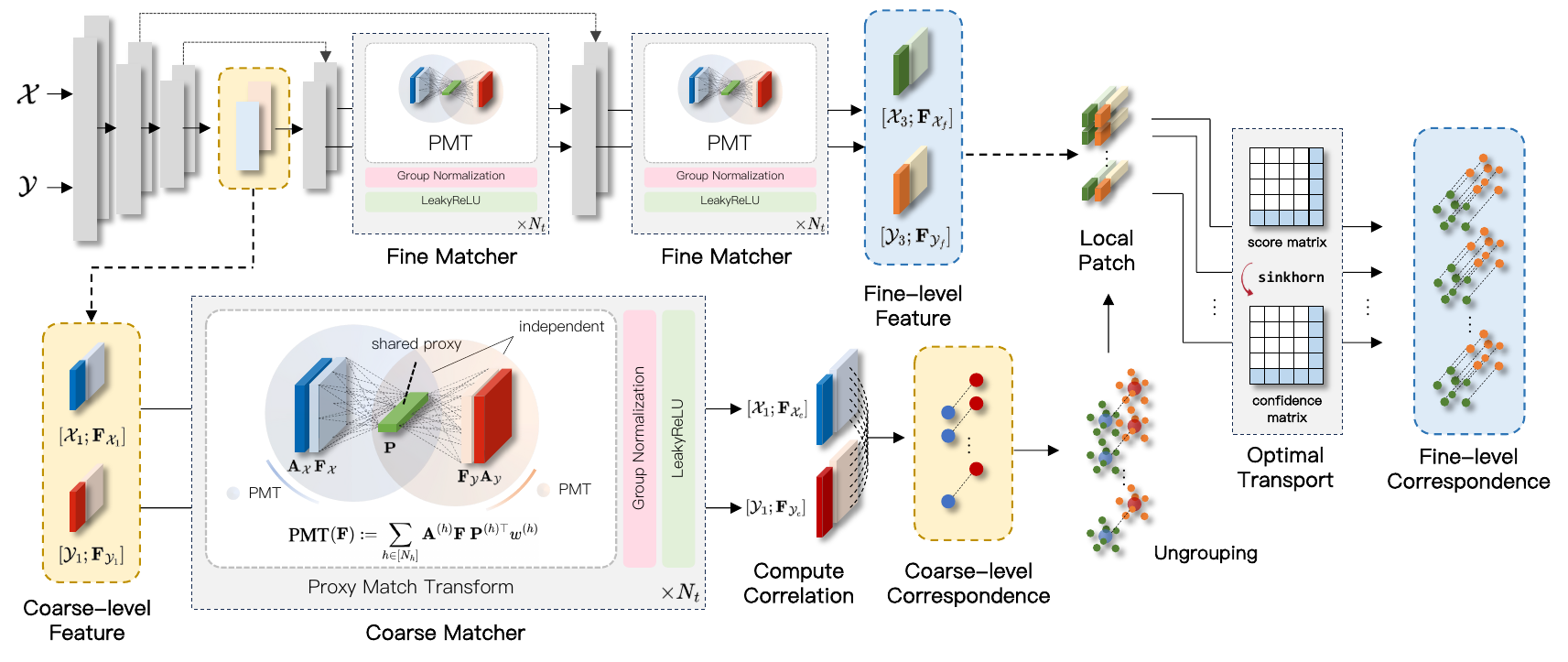}
\vspace{-8mm}
\caption{Overall pipeline of the Proxy Match TransformeR (PMTR) for pairwise shape assembly. The proposed architecture consists of coarse-level matching and fine-level matching. Each part of matching uses coarse-level features and fine-level features, respectively, acquired from the KPConv-FPN backbone as their input. Each matcher consists of $N_t$ PMT layers in series. See Sec.~\ref{sec:overall} for details.}
\vspace{-7mm}
\label{fig:architecture}
\end{center}
\end{figure*}

\subsection{Constraints for Proxy Match Transform}
\label{sec:constraint}

For the Proxy Match Transforms to express the high-order convolution, we assume the following constraints, (i) orthonormality constraint: $\mathbf{P}^{(i)\top} \mathbf{P}^{(j)} = \mathbf{I}_{D_{\text{emb}}}$ if $i=j$, and (ii) zero-matrix constraint: $\mathbf{P}^{(i)\top} \mathbf{P}^{(j)} = \mathbf{0} \in \mathbb{R}^{D_{\text{emb}} \times D_{\text{emb}}}$ otherwise for all $i, j \in [N_h]$.
Under such conditions, a dot product between two Proxy Match Transforms can effectively approximate high-order convolution.
Our main theoretical result is provided below.

\smallbreak
\noindent \textbf{Theorem 1.} {\em If we assume $\mathbf{P}^{(i)\top} \mathbf{P}^{(j)} = \mathbf{I}_{D_{\text{emb}}}$ if $i=j$ and $\mathbf{P}^{(i)\top} \mathbf{P}^{(j)} = \mathbf{0}$ otherwise for all $i, j \in [N_h]$, and define ${\mathbf{A}}^{(h)}_{(\mathbf{x}, \mathbf{y}), (\mathbf{n}, \mathbf{m})} \coloneqq {{\mathbf{A}_{\mathcal{X}}^{(h)}}}_{(\mathbf{x}, \mathbf{n})} \cdot {\mathbf{A}_{\mathcal{Y}}^{(h)}}_{(\mathbf{y}, \mathbf{m})}$ and ${w}^{(h)} \coloneqq {w}_{\mathcal{X}}^{(h)} {w}_{\mathcal{Y}}^{(h)}$, then, the dot-product of Proxy Match Transform outputs with a sufficient number of heads $N_h$ can express high-dimensional convolutional layer with kernel $K: \mathbb{R}^{6} \xrightarrow{} \mathbb{R}$: $\text{PMT}(\mathbf{F}_{\mathcal{X}}) \cdot \text{PMT}(\mathbf{F}_{\mathcal{Y}})^{\top} = \text{Conv}(\mathbf{F}_{\mathcal{X}}, \mathbf{F}_{\mathcal{Y}})$.}

\smallbreak
We refer to the Appendix~\ref{sec:theorem} for the complete proof of the theorem.
For the proxy tensors to satisfy the conditions, we design two auxiliary training objectives on proxy tensors, orthonormal loss $\mathcal{L}_{\text{orth}}$ and zero loss $\mathcal{L}_{\text{zero}}$, as follows:
\begin{align}
    \mathcal{L}_{\text{orth}} = \sum_{(i,j)\in [N_{h}]^{2}} \delta(i, j) (\mathbf{P}^{(i)\top} \mathbf{P}^{(j)} - \mathbf{I}_{D_{\text{emb}}}), \\
    \mathcal{L}_{\text{zero}} = \sum_{(i,j)\in [N_{h}]^{2}} (1-\delta(i,j)) \mathbf{P}^{(i)\top} \mathbf{P}^{(j)},
    \label{eq:aux_loss}
\end{align}
where $\delta(i, j)$ provides 1 if $i = j$ and 0 otherwise.

\subsection{Overall architecture}
\label{sec:overall}
The proposed architecture, dubbed Proxy Match TransformeR (PMTR) comprises four main parts:
(1) feature extraction, (2) coarse-level matching, (3) fine-level matching, and (4) transformation prediction \& training objectives.
As illustrated in Fig.~\ref{fig:architecture}, our pipeline begins with the point cloud pair embedding.
The feature extraction network generates three pairs of features, each at distinct spatial resolutions.
These feature pairs are subsequently fed to a corresponding PMT layer, which facilitates both coarse-level matching (for mating surface localization) and fine-level matching (for geometric matching). 
The outputs from the coarse matching phase are utilized to establish a preliminary correspondence between the mating surfaces of the input parts, which is crucial for identifying potential areas of alignment.
Subsequently, the fine matching phase is designed to refine these correspondences, focusing exclusively on reliable matches identified during the coarse matching stage.
This allows for precise correspondence establishment, ensuring accurate assembly as demonstrated by our experiments in Sec.~\ref{sec:ablation}.

\textbf{Feature extraction.} 
A pair of point clouds to match is fed to a feature embedding network, reducing their spatial resolution to provide a coarse-level feature pair.
Each of the two subsequent upsampling layers connected in series provides features in a higher resolution.
From this U-Net-shaped architecture, similar to KPConv-FPN~\citep{thomas2019kpconv}, the model gives three pairs of point cloud features with different spatial resolutions:
$\{(\mathbf{F}_{\mathcal{X}_{n}}, \mathbf{F}_{\mathcal{Y}_{n}})\}_{n=1}^{3}$ where $\mathbf{F}_{\mathcal{X}_{n}} \in \mathbb{R}^{|\mathcal{X}_{n}| \times D_{n\text{-emb}}}$ with $|\mathcal{X}_{1}| < |\mathcal{X}_{2}| < |\mathcal{X}_{3}|$, which implies that $\mathbf{F}_{\mathcal{X}_{1}}$ is the coarse feature with the smallest number of features.
$\{ \mathcal{Y}_{i} \}_{n=1}^{3}$ is similarly defined.
The coarse feature pair $\{(\mathbf{F}_{\mathcal{X}_{1}}, \mathbf{F}_{\mathcal{Y}_{1}})\}$ is used to identify potential mating surfaces to match, while the others $\{(\mathbf{F}_{\mathcal{X}_{n}}, \mathbf{F}_{\mathcal{Y}_{n}})\}_{n=2}^{3}$ are used to precisely align the identified potential surface matches.

\textbf{Coarse-level matching.} 
At this stage, PMT processes the coarse feature pair $\{(\mathbf{F}_{\mathcal{X}_{1}}, \mathbf{F}_{\mathcal{Y}_{1}})\}$, in order to evaluate the potential local correspondence between the feature set.
This is achieved without directly computing the pairwise correlation matrix $\mathbf{F}_{\mathcal{X}_{1}} \cdot {\mathbf{F}_{\mathcal{Y}_{1}}}^{\top}$, which would otherwise result in a quadratic dimensionality of $\mathbb{R}^{|\mathcal{X}_{1}| \times |\mathcal{Y}_{1}|}$.
Instead, a pair of PMTs operates in a manner that allows them to be refined independently to provide two refined coarse-level features $(\mathbf{F}_{\mathcal{X}_{c}}, \mathbf{F}_{\mathcal{Y}_{c}})$ as follows:
\begin{align}
    \text{PMT}(\mathbf{F}_{\mathcal{X}_{1}}) = \mathbf{F}_{\mathcal{X}_{c}}, \ \ \ \ \
    \text{PMT}(\mathbf{F}_{\mathcal{Y}_{1}}) = \mathbf{F}_{\mathcal{Y}_{c}}.
\end{align}
Despite this independence, the transformations ensure that the dot product of the refined features closely approximates the output of a high-order feature transformation.
The approximation is conceptualized as if the features had undergone a high-order transform, according to Theorem 1:
\begin{align}
    \text{PMT}(\mathbf{F}_{\mathcal{X}_{1}}) \cdot \text{PMT}(\mathbf{F}_{\mathcal{Y}_{1}})^{\top} &= \mathbf{F}_{\mathcal{X}_{c}} \cdot {\mathbf{F}_{\mathcal{Y}_{c}}}^{\top} \\
    &\approx \text{Conv}(\mathbf{F}_{\mathcal{X}_{1}}, \mathbf{F}_{\mathcal{Y}_{1}}).
\end{align}
This approach allows for independent refinement of the features while still capturing the essence of their interaction, akin to high-order convolution {\em without the direct computation of their pairwise correlation}, thereby effectively avoiding the burden of quadratic complexity.

\textbf{Fine-level matching.}
In fine-level matching, we leverage the high-resolution feature pairs $\{(\mathbf{F}_{\mathcal{X}_{n}}, \mathbf{F}_{\mathcal{Y}_{n}})\}_{n=2}^{3}$ to achieve a more precise alignment.
This stage mirrors coarse-level matching in its use of PMT layers for transforming features but in a serial configuration.
This setup ensures that the output of one PMT layer serves as the input to the next, such that $\text{PMT}(\mathbf{F}_{\mathcal{X}_{2}}) = \mathbf{F}_{\mathcal{X}_{3}}$ and, subsequently, $\text{PMT}(\mathbf{F}_{\mathcal{X}_{3}}) = 
\mathbf{F}_{\mathcal{X}_{f}}$, with an analogous sequence to provide fine-level features $\mathbf{F}_{\mathcal{Y}_{f}}$.
Note that PMT effectively addresses the infeasibility of employing vanilla high-order convolution for high-resolution matching, especially under conditions where $|\mathcal{X}|, |\mathcal{Y}| > 1500$, as observed in our experiments.
Compared to vanilla high-order convolution with complexity of $\mathcal{O}(|\mathcal{X}| \cdot |\mathcal{Y}|)$, rendering it infeasible for large-scale applications, the proposed PMT having $\mathcal{O}(\text{max}(|\mathcal{X}|, |\mathcal{Y}|) \cdot D_{\text{proxy}})$ complexity where $D_{\text{proxy}} \ll |\mathcal{X}|,|\mathcal{Y}|$ provides a more efficient means of analyzing feature correlations.
In Sec.~\ref{sec:ablation}, we present apples-to-apples comparisons, illustrating the practical advantages of PMT over traditional matching methods, \eg, high-order convolution~\citep{rocco2018neighbourhood}.

\begin{table*}[!t]
    \centering
    \caption{Pairwise shape assembly results. Numbers in bold indicate the best performance and underlined ones are the second best.}
    \label{table:everyday_mating}
    \vskip 0.05in
    \resizebox{\textwidth}{!}{
	\begin{tabular}{l|cc|rrrr|rrrr}
        \toprule
        \multirow{3.5}{*}{Method} & \multirow{3.5}{*}{\makecell{Estimator Type}} & \multirow{3.5}{*}{\makecell{Target \\ $\{\mathbf{R}|\mathbf{t}\}$}}  & CRD $\downarrow$ & CD $\downarrow$ & RMSE (R) $\downarrow$ & RMSE (T) $\downarrow$ & CRD $\downarrow$ & CD $\downarrow$ & RMSE (R) $\downarrow$ & RMSE (T) $\downarrow$ \\
        
        & & & ($10^{-2}$) & ($10^{-3}$) & ($^{\circ}$) & ($10^{-2}$) & ($10^{-2}$) & ($10^{-3}$) & ($^{\circ}$) & ($10^{-2}$) \\
        \cmidrule{4-11}

        & & & \multicolumn{4}{c|}{\texttt{everyday}} & \multicolumn{4}{c}{\texttt{artifact}} \\
        \midrule
        
        Global~\yrcite{schor2019componet, li2020global} %
            & \multirow{5}{*}{\makecell{MLP}} & \multirow{5}{*}{\makecell{absolute\\pose}}
            & 27.77 & 15.26 & 110.74 & 30.61 & 19.26 & 7.16 & 86.30 & 21.02 \\
        LSTM~\yrcite{wu2020lstm} %
            & & & 20.04 & 7.77 & 84.60 & 22.07 & 19.52 & 6.45 & 84.42 & 21.33 \\
        DGL~\yrcite{huang2020dgl} %
            & & & 20.32 & 6.40 & 86.23 & 22.38 & 19.82 & 6.19 & 85.46 & 21.65 \\
        NSM~\yrcite{chen2022neural} %
            & & & 21.71 & 11.09 & 83.38 & 23.71 & 19.44 & 6.33 & 83.22 & 21.41 \\
        \citet{wu2023leveraging}  %
            & & & 20.65 & 11.66 & 84.58 & 22.90 & 19.17 & 7.97 & 85.04 & 20.90 \\
        \midrule
        GeoTransformer~\yrcite{qin2022geotransformer} %
            & \multirow{3}{*}{\makecell{correspondence\\alignment}} & \multirow{3}{*}{\makecell{relative\\transformation}}
             & \underline{0.61} & \underline{0.51} & \underline{22.81} & 7.28 & \underline{0.89} & \underline{0.70} & \underline{33.23} & 10.30 \\
        Jigsaw~\yrcite{lu2023jigsaw}  %
            & & & 5.48 & 1.34 & 38.73 & \textbf{2.73} & 6.36 & 1.45 & 39.71 & \textbf{3.02} \\
        \textbf{PMTR (Ours)} %
            & & & \textbf{0.39} & \textbf{0.25} & \textbf{17.14} & \underline{5.53} & \textbf{0.60} & \textbf{0.42} & \textbf{23.28} & \underline{7.27} \\
        \bottomrule
	\end{tabular}
    }
\end{table*}

\textbf{Transformation prediction.} 
After the coarse-level matching, the refined feature pair $(\mathbf{F}_{\mathcal{X}_{c}}, \mathbf{F}_{\mathcal{Y}_{c}})$ is utilized to compute correlation in size of $|\mathcal{X}_{c}| \times |\mathcal{Y}_{c}|$ where each score at position $(\mathbf{x}, \mathbf{y})$ is defined as $\text{exp}(-||(\mathbf{F}_{\mathcal{X}_{c}})_{\mathbf{x}}-(\mathbf{F}_{\mathcal{Y}_{c}})_{\mathbf{y}}||^2_2)$ similarly to the work of~\citet{qin2022geotransformer}.
From $|\mathcal{X}_{c}| \times |\mathcal{Y}_{c}|$ number of scores, we collect top-$k$ matches as reliable coarse matches, laying the foundation for more granular alignment at the subsequent fine-level matching.
Building on coarse-level {\em matches} and fine-level {\em features}, we employ the point-to-node grouping method~\citep{yu2021cofinet}, which clusters fine-level features that are spatially proximate to the coarse matches, effectively sharpening the broad coarse-level correspondence into more precise fine-level ones.
In essence, the computation of fine-level matches is directly influenced by the groundwork laid at the coarse level, establishing a hierarchical refinement process.
We then incorporate an optimal transport layer~\citep{sarlin2020superglue} into the fine-level matches to obtain final correspondences for the subsequent transformation prediction.
Finally, similarly to~\citet{qin2022geotransformer}, we use the final correspondences to predict the relative transformation $\{\mathbf{R}|\mathbf{t}\}$ between the point cloud pair $(\mathcal{X}, \mathcal{Y})$.

\textbf{Training objectives.} 
Following the previous 3D matching literatures~\citep{zhao2023learning,wu2023graph,Chen_2023_ICCV,Yu_2023_CVPR}, we adopt overlap-aware circle loss $\mathcal{L}_{\text{oc}}$~\citep{qin2022geotransformer}, and point matching loss $\mathcal{L}_{\text{p}}$~\citep{sarlin2020superglue}, as our main training objectives for coarse- and fine-level correspondence matching, respectively.
We direct the reader to the work of \citet{qin2022geotransformer} for more details of $\mathcal{L}_{\text{oc}}$ and $\mathcal{L}_{\text{p}}$.
With two auxiliary losses in Eq.~\ref{eq:aux_loss}, our main training objective is defined as follows:
\begin{align}
    \mathcal{L} = {\mathcal{L}_{\text{oc}} + \mathcal{L}_{\text{p}}} + \lambda_{\text{orth}}\mathcal{L}_{\text{orth}} + \lambda_{\text{zero}}\mathcal{L}_{\text{zero}},
\end{align}
where $\lambda_{\text{orth}}$ and $\lambda_{\text{zero}}$ are hyperparameters which are set to 1.0 in our experiments.



\section{Experiments}

In this section, we discuss the dataset and evaluation metrics used (Sec.~\ref{sec:dataset}), implementation details (Sec.~\ref{sec:impl}), the results of pairwise shape assembly with comprehensive analysis (Sec.~\ref{sec:result}), an in-depth ablation study to inspect the efficacy of the proposed techniques (Sec.~\ref{sec:ablation}), and extension of our evaluation to the task of multi-part assembly (Sec.~\ref{sec:mpa}).

\subsection{Dataset and Evaluation Metrics}
\label{sec:dataset}

\smallbreak
\noindent \textbf{Dataset.} In our experiments, we utilize the Breaking Bad dataset~\citep{sellan2022breakingbad}, a large-scale dataset of fractured objects for the task of geometric shape assembly, which consists of over 1 million fractured objects simulated from 10K meshes of PartNet~\citep{mo2019partnet} and Thingi10k~\citep{zhou2016thingi10k}.
For {\em pairwise} assembly training and evaluation, we exclusively select a subset of the Breaking Bad dataset that contains two-part objects (Sect.~\ref{sec:result}).
For {\em multi-part} assembly, we expand our evaluation to include all samples in the dataset, encompassing objects with 2 to 20 parts (Sec.~\ref{sec:mpa}).

\smallbreak
\noindent \textbf{Evaluation metrics.}
Following the evaluation protocol of ~\citet{sellan2022breakingbad}, we measure the root mean square error (RMSE) between the ground-truth and predicted rotation (R) and translation (T) parameters, and the Chamfer distance (CD) between the assembly results and ground-truth. In addition, we introduce and report a new metric, called CoRrespondence Distance (CRD), which is defined as the Frobenius norm between the input pair of the assembled point cloud;
unlike CD, CRD offers a more comprehensive measure of correspondence, capturing both proximity and structural alignment between the assembled objects.
We compute the evaluation metrics of RMSE (R) and RMSE (T) based on \textit{relative transformation}, \eg, rotation and translation, between the input fracture pair, instead of the absolute pose as in the previous literature~\citep{chen2022neural, wu2023leveraging} by setting the largest fracture as an anchor and computing the relative transformation.
The formal definitions of the evaluation metrics can be found in the Appendix~\ref{sec:metric}.

\subsection{Implementation details}
\label{sec:impl}
We implement our PMTR using PyTorch Lightning~\citep{falcon2019lightning}. 
The experiments were carried out on a machine with Intel(R) Xeon(R) Gold 6342 CPU @ 2.80GHz and NVIDIA GeForce RTX 3090 GPU. 
For all experiments, except those including GeoTransformer, we use ADAM~\citep{kingma2015adam} optimizer with a learning rate of $1\times10^{-3}$ for 150 epochs. 
For GeoTransformer, we use identical settings but only reduce the learning rate to $1\times10^{-4}$ to prevent model divergence. 
To ensure uniform point density among fractures, we uniform-sample approximately 5,000 points on the surface of holistic objects and allocate the number of sample points for each fracture proportional to the surface area of each fracture.
Each of the coarse- and fine-level matchers consists of 2 PMT($\cdot$) layers ($N_t = 2$) with non-linearity and group norm~\citep{wu2018groupnorm}.
See Appendix~\ref{sec:implementation} for further details. 

\smallbreak
\noindent \textbf{Avoiding quadratic complexity of attention in PMT.}
In our actual implementation, we use local, \ie, sparse, attention for $\mathbf{A}_{\mathcal{X}}^{(h)}$ by collecting attention scores of the `neighborhood' of each position, thus reducing attention size to ${|\mathcal{X}| \times \epsilon}$ instead of ${|\mathcal{X}| \times |\mathcal{X}|}$ where $\epsilon \in \mathbb{N}^{+}$ is the number of neighbors: $\epsilon \ll |\mathcal{X}|$. 
Specifically, attention at position $\mathbf{x} \in \mathbb{R}^{3}$ denoted as $\mathbf{A}_{\mathcal{X} (\mathbf{x}, :)}^{(h)} \in \mathbb{R}^{1 \times \epsilon}$ is limited to the neighborhood of $\mathbf{x}$, represented by $\mathcal{N}(\mathbf{x})$.
Then, the output of PMT at $\mathbf{x}$ is formulated as $\text{PMT}(\mathbf{F}_{\mathcal{X}})_{(\mathbf{x}, :)} = \sum_{h \in [N_h]} \mathbf{A}_{\mathcal{X} (\mathbf{x}, :)}^{(h)} \mathbf{F}_{\mathcal{X} (\mathcal{N}(\mathbf{x}),:)} \mathbf{P}^{(h)\top} w_{\mathcal{X}}^{(h)}$ where $\mathbf{F}_{\mathcal{X} (\mathcal{N}(\mathbf{x}),:)} \in \mathbb{R}^{\epsilon \times D_{\text{emb}}}$ represents the neighborhood features of position $\mathbf{x}$.
This method significantly reduces the computational complexity typically associated with full pairwise attention, which would otherwise be quadratic, \ie, $|\mathcal{X}| \times |\mathcal{X}|$. This reduction in complexity mirrors the strategies found in the existing literature, such as those described by~\citet{thomas2019kpconv}.
However, for the sake of simplicity, we narrate with a conventional form of square attention matrices $\mathbf{A}_{\mathcal{X}}^{(h)} \in \mathbb{R}^{|\mathcal{X}| \times |\mathcal{X}|}$ or $\mathbf{A}_{\mathcal{Y}}^{(h)} \in \mathbb{R}^{|\mathcal{Y}| \times |\mathcal{Y}|}$ to illustrate our methodology in this paper. 

\noindent\textbf{Assessment with relative transformations.}
Note that the previous methods for pairwise geometric assembly~\citep{li2020global,wu2020lstm,huang2020dgl,chen2022neural} predict {\em two different transformation parameters} for the input pair of parts to assemble them in 3D space.
However, this approach has a limitation in the evaluation:
although a model perfectly assembles the pair of parts, the assessment may be inaccurate if the assembled object does not match the specific {\em absolute} pose of the ground truth.
To address this issue, we suggest predicting the {\em relative} transformation between the input parts, allowing us to focus solely on the assembly rather than the predefined absolute poses.

\subsection{Pairwise Shape Assembly}

To evaluate our method, we categorize the previous methods into two groups based on their approach to transformation parameters $\{\mathbf{R}|\mathbf{t}\}$ prediction.
The first group includes `regression methods' that encode each part into a global embedding and directly regress their absolute transformations using MLP:
Global~\citep{li2020global}, LSTM~\citep{wu2020lstm}, DGL~\citep{huang2020dgl}, NSM~\citep{chen2022neural}, and \citet{wu2023leveraging}.
The second group consists of `matching-based methods' that estimate relative transformations by aligning their predicted correspondences between each pair of parts: GeoTransformer~\citep{qin2022geotransformer} and Jigsaw~\citep{lu2023jigsaw}.

\begin{figure*}[ht]
\begin{center}
\includegraphics[width=\textwidth]{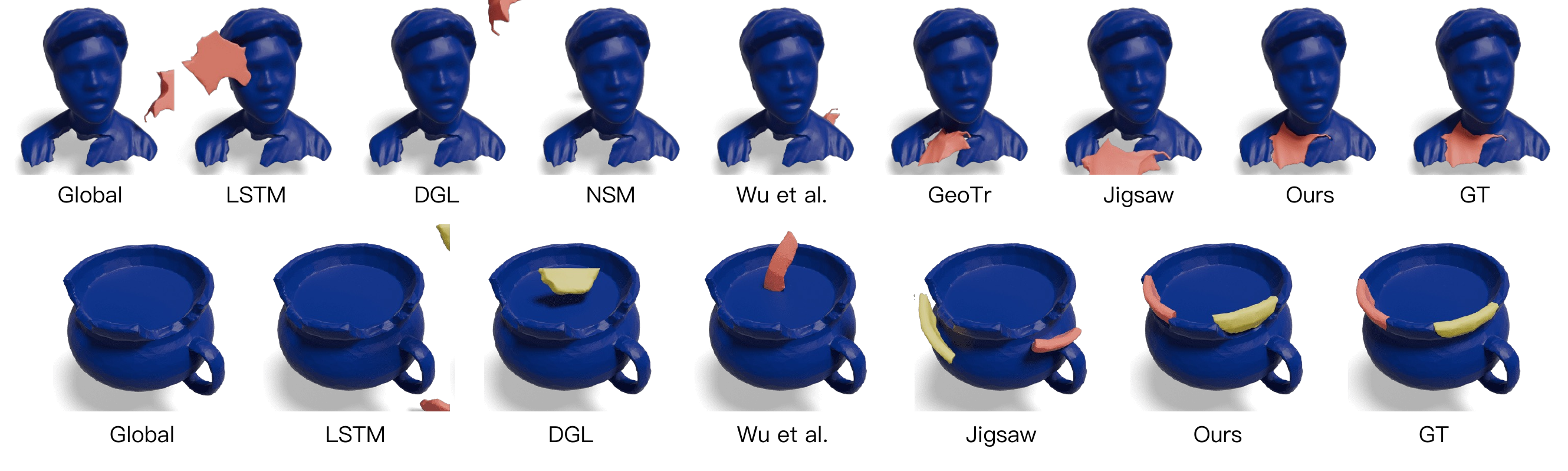}
\vspace{-8mm}
\caption{Qualitative results of pairwise shape assembly (Upper row) and multipart shape assembly (Bottom row) on Breaking Bad dataset.}
\label{fig:qual_main}
\end{center}
\vskip -0.2in
\end{figure*}

\smallbreak
\noindent\textbf{Experimental results and analysis.}
\label{sec:result}
We evaluate our method and compare it against baseline methods on the \texttt{everyday} and \texttt{artifact} subsets of the Breaking Bad dataset. 
Tab.~\ref{table:everyday_mating} presents the results, which demonstrate that our method consistently outperforms all baseline methods on both subsets.
In Fig.~\ref{fig:qual_main}, we provide qualitative comparisons between ours and the baselines, using mesh representation for better visualization.

\begin{figure}[b]
   \centering
   \begin{tabular}{cc}
   \includegraphics[width=0.45\linewidth]{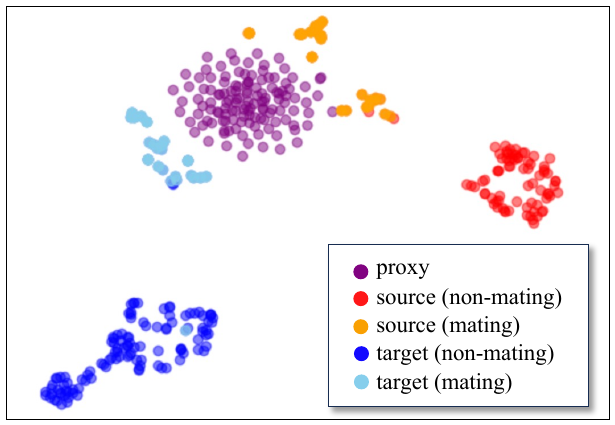} &
   \includegraphics[width=0.45\linewidth]{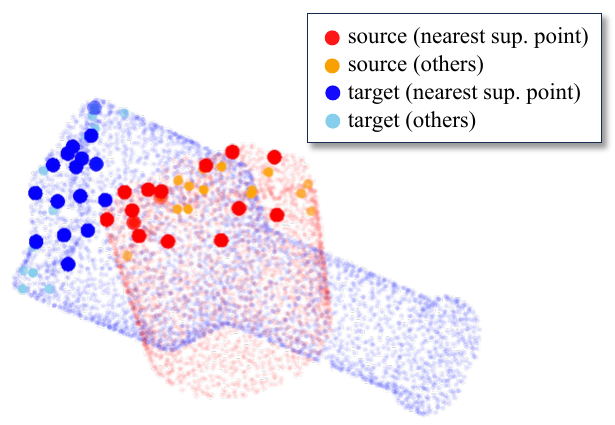} \\
   (a) & (b)
   \end{tabular}
   \vskip -0.05in
\caption{(a) t-SNE visualization of proxy tensor (colored in purple), source features $\mathbf{F}_{\mathcal{X}_{c}}$ and target features $\mathbf{F}_{\mathcal{Y}_{c}}$. The source and target features are colored in warm (red) and cool (blue) tones, respectively, and those on mating surfaces are colored in orange and lightblue. (b) Feature visualization in 3D space. Source $\mathcal{X}_{1}$ and target features $\mathcal{Y}_{1}$ with closer proximity to the proxy tensor are highlighted in red and blue, respectively, and features on mating surfaces are highlighted in orange and lightblue. For this visualization, we use a proxy tensor at a head index of $h=0$: $\mathbf{P}^{(0)}$.}
\label{fig:proxy_vis}
\vspace{-5mm}
\end{figure}

To provide deeper insights into the learned shared proxy $\mathbf{P}^{(h)}$, we visualize how the proxy and the refined coarse-level features ($\mathbf{F}_{\mathcal{X}_{c}}$ and $\mathbf{F}_{\mathcal{Y}_{c}}$) are distributed in the feature space via t-SNE.
As shown in Fig.~\ref{fig:proxy_vis}, the visualization reveals that subsets of the source and target features (orange and lightblue) and the proxy (purple) form distinct clusters (Fig.~\ref{fig:proxy_vis} (a)) with closer proximity, implying higher correlation with the proxy.
In Fig.~\ref{fig:proxy_vis}. (b), we visually mark those points in the 3D space.
Notably, the points with the highest correlation (red and blue) with the proxy are predominantly located on the ``mating surfaces'' of the parts, revealing that the proxy effectively facilitates the information exchange between the given features without the burden of quadratic complexity.

\begin{table}[t]
    \centering
    \caption{
		\textbf{Ablation study on the proxy sharing.} By sharing proxy tensor in each Proxy Match Transform layer, two independent feature transforms share information, yielding the highest score.
	}
        \vskip 0.05in
	\label{table:ablation_proxy}
	\resizebox{.99\linewidth}{!}{
	\begin{tabular}{ccc|cccc}
        \toprule
        \multirow{2}{*}{Ref.} &  \multirow{2}{*}{proxy} & shared & CRD $\downarrow$ & CD $\downarrow$ & RMSE (R) $\downarrow$ & RMSE (T) $\downarrow$ \\
                &  & proxy & ($10^{-2}$) & ($10^{-3}$) & ($^{\circ}$) & ($10^{-2}$) \\
        
        \midrule 
        (a) &  \textcolor{red}{\ding{55}} & \textcolor{red}{\ding{55}} & 0.53 & 0.47 & 21.04 & 6.93 \\
        (b) & \textcolor{green}{\ding{51}} & \textcolor{red}{\ding{55}} & \underline{0.44} & \underline{0.31} & \underline{18.66} & \underline{5.97} \\
        \textbf{Ours} & \textcolor{green}{\ding{51}} & \textcolor{green}{\ding{51}} & \textbf{0.39} & \textbf{0.25} & \textbf{17.14} & \textbf{5.53} \\

        \bottomrule
        
	\end{tabular}
	}
\vspace{-4mm}
\end{table}
\begin{table}[t]
    \centering
    \caption{
  \textbf{Ablation study on the contribution of $\mathcal{L}_{\textnormal{orth}}$ and $\mathcal{L}_{\textnormal{zero}}$.} They constrain Proxy Match Transform in approximating the high-dimensional convolution layers, yielding the highest score.}
        \vskip 0.05in
	\label{table:ablation_loss}
	\resizebox{.99\linewidth}{!}{
	\begin{tabular}{ccc|cccc}
        \toprule
        \multirow{2}{*}{Ref.} & \multirow{2}{*}{$\mathcal{L}_{\text{orth}}$} & \multirow{2}{*}{$\mathcal{L}_{\text{zero}}$} & CRD $\downarrow$ & CD $\downarrow$ & RMSE (R) $\downarrow$ & RMSE (T) $\downarrow$ \\
        &  &  & ($10^{-2}$) & ($10^{-3}$) & ($^{\circ}$) & ($10^{-2}$) \\
        
        \midrule
        (a) & \textcolor{red}{\ding{55}} & \textcolor{red}{\ding{55}} & 0.43 & 0.31 & 18.82 & 6.23 \\
        (b) & \textcolor{green}{\ding{51}} & \textcolor{red}{\ding{55}} & \underline{0.43} & 0.32 & \underline{17.87} & \underline{5.73} \\
        (c) & \textcolor{red}{\ding{55}} & \textcolor{green}{\ding{51}} & 0.43 & \underline{0.27} & 18.77 & 6.02 \\
        \textbf{Ours} & \textcolor{green}{\ding{51}} & \textcolor{green}{\ding{51}} & \textbf{0.39} & \textbf{0.25} & \textbf{17.14} & \textbf{5.53} \\

        \bottomrule
        
	\end{tabular}
	}
\vspace{-4mm}
\end{table}
\begin{table}[t]
    \centering
    \caption{
        \textbf{Ablation study on the choice of fine-level matcher.} Proxy Match Transform layer at fine-level yields the best assembly accuracy while incurring low-compute complexity than baselines.
	}
        \vskip 0.05in
	\label{table:ablation_finematcher_1}
	\resizebox{\linewidth}{!}{
	\begin{tabular}{ccc|cccc}
        \toprule
        \multirow{2}{*}{Ref.} & Coarse-level & Fine-level & CRD $\downarrow$ & CD $\downarrow$  & RMSE (R) $\downarrow$  & RMSE (T) $\downarrow$  \\
        & Matcher & Matcher & ($10^{-2}$) & ($10^{-3}$) & ($^{\circ}$) & ($10^{-2}$) \\
        \midrule
        (a) %
            & \multirow{6}{*}{PMT} & None & 0.53 & 0.43 & 20.70 & 6.63 \\
        (b) %
            & & Linear & \underline{0.47} & \underline{0.37} & 17.55 & \underline{5.68} \\
        (c) %
            & & MLP & 0.49 & 0.38 & \underline{17.35} & 5.69 \\
        (d) %
            & & HDC & \multicolumn{4}{c}{\cellcolor{gray!25}Out of memory error} \\
        (e) %
            & & GeoTr & \multicolumn{4}{c}{\cellcolor{gray!25}Out of memory error} \\
        \textbf{Ours} %
            & & PMT & \textbf{0.39} & \textbf{0.25} & \textbf{17.14} & \textbf{5.53} \\

        \bottomrule
        
	\end{tabular}
	}
\vspace{-4mm}
\end{table}
\begin{table}[t]
    \centering
    \caption{
        \textbf{Ablation study on the impact of Proxy Match Transform as a fine-level matcher.} Proxy Match Transform layer consistently boosts performance with various coarse-level matchers.
	}
        \vskip 0.05in
	\label{table:ablation_finematcher_2}
	\resizebox{\linewidth}{!}{
	\begin{tabular}{ccc|cccc}
		\toprule
        
        \multirow{2}{*}{Ref.} & Coarse-level & Fine-level & CRD $\downarrow$ & CD $\downarrow$  & RMSE (R) $\downarrow$  & RMSE (T) $\downarrow$  \\
        & Matcher & Matcher & ($10^{-2}$) & ($10^{-3}$) & ($^{\circ}$) & ($10^{-2}$) \\
        \midrule
        (a) %
            & \multirow{2}{*}{None} & None & 0.69 & 0.57 & 27.71 & 8.78 \\
        (b) %
            & & PMT & \textbf{0.60} & \textbf{0.52} & \textbf{24.66} & \textbf{7.42} \\
            
        \midrule
        (c) %
            & \multirow{2}{*}{Linear} & None & 0.64 & 0.53 & 26.14 & 7.42 \\
        (d) %
            & & PMT & \textbf{0.55} & \textbf{0.50} & \textbf{22.44} & \textbf{6.69} \\

        \midrule
        (e) %
            & \multirow{2}{*}{MLP} & None & 0.66 & 0.50 & 26.93 & 7.35 \\
        (f) %
            & & PMT & \textbf{0.57} & \textbf{0.44} & \textbf{23.74} & \textbf{7.03} \\

        \midrule
        (g) %
            & \multirow{2}{*}{HDC} & None & 0.76 & 0.63 & 27.75 & 8.68 \\
        (h) %
            & & PMT & \textbf{0.63} & \textbf{0.48} & \textbf{23.43} & \textbf{7.08} \\

        \midrule
        (i) %
            & \multirow{2}{*}{GeoTr} & None & 0.61 & 0.51 & \textbf{22.81} & \textbf{7.28} \\
        (j) %
            & & PMT & \textbf{0.48} & \textbf{0.33} & 23.91 & 7.32 \\

        \midrule
        (k) %
            & \multirow{2}{*}{PMT} & None & 0.53 & 0.43 & 20.70 & 6.63 \\
        \textbf{Ours} %
            & & PMT & \textbf{0.39} & \textbf{0.25} & \textbf{17.14} & \textbf{5.53} \\

        \bottomrule
        
	\end{tabular}
	}
\vspace{-5mm}
\end{table}

\subsection{Ablation studies}
\label{sec:ablation}

\smallbreak
\noindent\textbf{Effect of proxy tensor in assembly.}
To verify the effect of proxy tensor in PMT, we conducted a series of ablation studies on the \texttt{everyday} subset of the Breaking Bad dataset.
Specifically, we examine the impact of the shared proxy tensor by either removing it or using two different proxies instead of a shared one.
The results, summarized in Tab.~\ref{table:ablation_proxy}, clearly indicate that both removing the proxy and not sharing it lead to a significant decline in assembly performance.
This underscores the efficacy of the shared proxy in facilitating information exchange in PMT.

Next, we explore the impact of $\mathcal{L}_{\text{orth}}$ and $\mathcal{L}_{\text{zero}}$, which serve as sufficient conditions to constrain the PMT layer to represent the high-dimensional convolutional layers, as detailed in Sec.~\ref{sec:constraint}. 
The results are presented in Tab.~\ref{table:ablation_loss};
as is evident from the table, the best performance is achieved when both losses are incorporated. 
This highlights the significance of these constraining conditions for PMT, as they are crucial in enabling PMT to effectively approximate the high-dimensional convolution.

\begin{table*}[ht]
    \centering
    \caption{Multi-part assembly results on the Breaking Bad dataset.}
    \vskip 0.05in
	\label{table:multipart}
	\resizebox{1.0\textwidth}{!}{
	\begin{tabular}{l|rrrrrr|rrrrrr}
        \toprule
        \multirow{3.5}{*}{Method} & CRD $\downarrow$ & CD $\downarrow$ & RMSE (R) $\downarrow$ & RMSE (T) $\downarrow$ & $\text{PA}_\text{CRD}$ $\uparrow$ & $\text{PA}_\text{CD}$ $\uparrow$ & CRD $\downarrow$ & CD $\downarrow$ & RMSE (R) $\downarrow$ & RMSE (T) $\downarrow$ & $\text{PA}_\text{CRD}$ $\uparrow$ & $\text{PA}_\text{CD}$ $\uparrow$ \\
        
        & ($10^{-2}$) & ($10^{-3}$) & (${}^{\circ}$) & ($10^{-2}$) & (\%) & (\%) & ($10^{-2}$) & ($10^{-3}$) & (${}^{\circ}$) & ($10^{-2}$) & (\%) & (\%) \\
        \cmidrule{2-13}

        & \multicolumn{6}{c|}{\texttt{everyday}} & \multicolumn{6}{c}{\texttt{artifact}} \\
        \midrule

        Global~\yrcite{schor2019componet, li2020global} %
            & 27.79 & 15.30 & 55.42 & 15.31 & 36.42 & 37.90 & 26.42 & 14.92 & 54.41 & 14.48 & 36.67 & 36.97 \\
        LSTM~\yrcite{wu2020lstm}%
            & 27.69 & 15.23 & 54.78 & 15.24 & 36.74 & 38.97 & 28.15 & 14.61 & 53.59 & 15.49 & 36.67 & 37.25 \\
        DGL~\yrcite{huang2020dgl} %
            & 27.90 & 13.23 & 55.76 & 15.33 & 36.99 & 39.70 & 27.48 & 13.91 & 54.66 & 15.10 & 36.66 & 37.40 \\
        \citet{wu2023leveraging} %
            & 28.18 & 19.70 & 54.98 & 15.59 & 35.66 & 36.28 & 26.02 & 15.81 & 54.35 & \underline{14.27} & 36.63 & 37.02 \\
        Jigsaw~\yrcite{lu2023jigsaw} %
            & \underline{14.13} & \underline{11.82} & \underline{41.12} & \underline{11.74} & \underline{52.48} & \underline{60.26} & \underline{16.10} & \underline{9.53} & \underline{42.01} & 17.47 & \underline{56.93} & \underline{65.58}  \\
        \textbf{PMTR (Ours)} %
            & \textbf{6.51} & \textbf{5.56} & \textbf{31.57} & \textbf{9.95} & \textbf{66.95} & \textbf{70.56} & \textbf{5.67} & \textbf{4.33} & \textbf{31.58} & \textbf{10.08} & \textbf{66.96} & \textbf{71.61} \\
        
        \bottomrule
	\end{tabular}
 	}
\vspace{-3mm}
\end{table*}

\smallbreak
\noindent\textbf{Comparison between different matchers.}
To demonstrate the efficacy and efficiency of the proposed matching layer, PMT, we conduct ablations by either removing it (None) or replacing it with different layers: a single linear transformation (Linear), multi-layer perceptron (MLP), high-dimensional convolution (HDC by~\citet{min2021hypercorrelation}), and GeoTransformer (GeoTr by~\citet{qin2022geotransformer}).
In Tab.~\ref{table:ablation_finematcher_1}, we compare ours with other layers at fine-level;
Undoubtedly, the layers without any information exchange between source and target features, \eg, None, Linear, and MLP, show dramatic drops in performance. 
While the matching layers of HDC and GeoTr cause out-of-memory-error due to their quadratic complexity, being unable to be incorporated at fine-level with large input spatial resolutions, the proposed PMT not only efficiently processes source and target features without memory burden but also effectively exchanges information between them via proxy tensor.
In Tab.~\ref{table:ablation_finematcher_2}, similar experiments are conducted at coarse-level.
As evident from the tables, incorporating the PMT layer as both fine and coarse matcher consistently leads to superior performance, affirming its superiority over the state-of-the-art matching layers~\citep{min2021hypercorrelation,qin2022geotransformer}.

\subsection{Multi-part Assembly}
\label{sec:mpa}
To assess the generalizability of our method, we extend our evaluation to include multiple input parts, \ie, multi-part assembly, which requires the model to understand the pairwise correspondence relationships among all input parts.
Utilizing the two-part assembly framework (Fig.~\ref{fig:architecture}), it begins with computing relative transformations between each pair of $P$ parts.
We then construct a {\em pose graph} wherein each node and factor respectively represent an individual part and the predicted relative transformation, \ie, pose, between two parts.
To optimize this pose graph for assembly, we employ a recent transformation averaging method detailed in the work of~\citet{dellaert2020shonan}.
After the optimization, we evaluate our method using the metrics from pairwise assembly, supplemented by Part Accuracy ($\text{PA}_{\text{CD}}$)~\citep{huang2020dgl} -- the percentage of parts with Chamfer Distance less than the predefined threshold of 0.01 -- as well as CRD-based Part Accuracy ($\text{PA}_{\text{CRD}}$) with 0.1 threshold.
As seen from Tab.~\ref{table:multipart} and Fig.~\ref{fig:qual_main}, our method significantly surpasses all baselines on all metrics on the multi-part assembly, demonstrating robust generalization to multiple input scenarios.
For details on the evaluation metrics, see Appendix~\ref{sec:metric}.


\section{Scope and Limitations}
Despite the advances in efficient point cloud matching and shape assembly, our method still faces several limitations.
First, the accuracy of our method can be compromised in scenarios with extremely low overlap between point clouds, which can hinder the identification of reliable correspondences.
Second, our method, like many others in the field, requires extensive training on domain-specific datasets to achieve optimal performance.
Third, while our experiments demonstrate the efficacy of PMT in shape assembly tasks, it has not been extensively tested across other potential applications such as robotics, manufacturing, digital artistry, or even restoration of ancient artifacts via more accurate and detailed part assembly.
Thus, the applicability of our approach beyond geometric shape assembly remains to be fully validated.
We leave this to future work.

\section{Conclusion}
We have introduced a low-complexity, high-order feature transform layer, Proxy Match Transform, designed for efficient approximation of traditional compute-intensive high-order feature transforms.
The significant performance improvements over the state of the arts with a lower computational cost indicate its real-world applicability from artifact reconstruction to manufacturing.
Although the proposed method has been applied exclusively to geometric shape assembly in this work, its significant improvements in various evaluation metrics reveal its potential for broad applications.

\section*{Acknowledgements}

This work was supported by IITP grants (RS-2022-II220290: Visual Intelligence for Space-Time Understanding and Generation (30\%), RS-2021-II212068: AI Innovation Hub (60\%), RS-2019-II191906: AI Graduate School Program at POSTECH (5\%), RS-2021-II211343: AI Graduate School Program at SNU: 2021-0-01343 (5\%)) funded by the Korea government.
\section*{Impact Statement}

The advancements in geometric shape assembly hold paramount potential across numerous fields, from archaeological artifact reconstruction to industrial manufacturing.
This research can also advance the field of robotics, digital artistry, augmented reality, and computer-aided design (CAD) via more accurate and robust shape assembly.

\bibliography{egbib}
\bibliographystyle{icml2024}

\newpage
\appendix
\onecolumn

\section{Theoretical Analysis of Proxy Match Transform}
\label{sec:theorem}

We now derive sufficient conditions such that Proxy Match Transform can express high-dimensional convolution. Our main theoretical result is given below.
\smallbreak
\noindent \textbf{Theorem 1.} {\em If we assume $\mathbf{P}^{(i)\top} \mathbf{P}^{(j)} = \mathbf{I}_{D_{\text{emb}}}$ if $i=j$ and $\mathbf{P}^{(i)\top} \mathbf{P}^{(j)} = \mathbf{0}$ otherwise for all $i, j \in [N_h]$, and define ${\mathbf{A}}^{(h)}_{(\mathbf{x}, \mathbf{y}), (\mathbf{n}, \mathbf{m})} \coloneqq {{\mathbf{A}_{\mathcal{X}}^{(h)}}}_{(\mathbf{x}, \mathbf{n})} \cdot {\mathbf{A}_{\mathcal{Y}}^{(h)}}_{(\mathbf{y}, \mathbf{m})}$ and ${w}^{(h)} \coloneqq {w}_{\mathcal{X}}^{(h)} {w}_{\mathcal{Y}}^{(h)}$, then, the dot-product of Proxy Match Transform outputs with a sufficient number of heads $N_h$ can express high-dimensional convolutional layer with kernel $K: \mathbb{R}^{6} \xrightarrow{} \mathbb{R}$: $\text{PMT}(\mathbf{F}_{\mathcal{X}}) \cdot \text{PMT}(\mathbf{F}_{\mathcal{Y}})^{\top} = \text{Conv}(\mathbf{F}_{\mathcal{X}}, \mathbf{F}_{\mathcal{Y}})$.}

{\em Proof.} \ We first take the dot-product of Proxy Match Transform outputs and simplify:
\begin{align}
    \text{PMT}(\mathbf{F}_{\mathcal{X}}) \cdot \text{PMT}(\mathbf{F}_{\mathcal{Y}})^{\top} &= \left( \sum_{h \in [{N_h}]} \mathbf{A}_{\mathcal{X}}^{(h)} \mathbf{F}_{\mathcal{X}} \mathbf{P}^{(h)\top} {w}_{\mathcal{X}}^{(h)} \right) \left( \sum_{h \in [{N_h}]} \mathbf{A}_{\mathcal{Y}}^{(h)} \mathbf{F}_{\mathcal{Y}} \mathbf{P}^{(h)\top} {w}_{\mathcal{Y}}^{(h)} \right)^{\top} \\
    &= \sum_{(i, j) \in [{N_h}]^2} {w}_{\mathcal{X}}^{(i)} \mathbf{A}_{\mathcal{X}}^{(i)} \mathbf{F}_{\mathcal{X}} \mathbf{P}^{(i)\top} \mathbf{P}^{(j)} {\mathbf{F}_{\mathcal{Y}}}^{\top} \mathbf{A}_{\mathcal{Y}}^{(j)\top} {w}_{\mathcal{Y}}^{(j)} \\
    &= \sum_{(i, j) \in [{N_h}]^2} \delta(i, j) \left( {w}_{\mathcal{X}}^{(i)} \mathbf{A}_{\mathcal{X}}^{(i)} \mathbf{F}_{\mathcal{X}} {\mathbf{F}_{\mathcal{Y}}}^{\top} \mathbf{A}_{\mathcal{Y}}^{(j)\top} {w}_{\mathcal{Y}}^{(j)} \right) \\
    &= \sum_{h \in [{N_h}]} {w}_{\mathcal{X}}^{(h)} \mathbf{A}_{\mathcal{X}}^{(h)} \mathbf{F}_{\mathcal{X}} {\mathbf{F}_{\mathcal{Y}}}^{\top}\mathbf{A}_{\mathcal{Y}}^{(h)\top} {w}_{\mathcal{Y}}^{(h)},
\end{align}
where $\delta(i, j)$ provides 1 if $i=j$ and $0$ otherwise.
Using definitions of ${\mathbf{A}}^{(h)} \in \mathbb{R}^{|\mathcal{X}||\mathcal{Y}| \times |\mathcal{X}||\mathcal{Y}|}$ and ${w}^{(h)} \in \mathbb{R}$, the output at a specific position $(\mathbf{x}, \mathbf{y}) \in \mathbb{R}^{6}$ is as follows:
\begin{align}
    ({\text{PMT}(\mathbf{F}_{\mathcal{X}}) \cdot \text{PMT}(\mathbf{F}_{\mathcal{Y}})^{\top}})_{(\mathbf{x}, \mathbf{y})} 
    &= \sum_{h \in [{N_h}]} {\mathbf{A}_{\mathcal{X}}^{(h)}}_{(\mathbf{x}, :)} \mathbf{F}_{\mathcal{X}} {\mathbf{F}_{\mathcal{Y}}}^{\top} {\mathbf{A}_{\mathcal{Y}}^{(h)\top}}_{(:, \mathbf{y})} {w}^{(h)} \\
    &= \sum_{h \in [{N_h}]} \sum_{(\mathbf{n}, \mathbf{m}) \in \mathcal{X} \times \mathcal{Y}} {\mathbf{A}_{\mathcal{X}}^{(h)}}_{(\mathbf{x}, \mathbf{n})} {\mathbf{F}_{\mathcal{X}}}_{(\mathbf{n}, :)} {{\mathbf{F}_{\mathcal{Y}}}^{\top}}_{(:, \mathbf{m})}{\mathbf{A}_{\mathcal{Y}}^{(h)\top}}_{(\mathbf{m}, \mathbf{y})} {w}^{(h)} \\
    &= \sum_{h \in [{N_h}]} \left( \sum_{(\mathbf{n}, \mathbf{m}) \in \mathcal{X} \times \mathcal{Y}} {\mathbf{A}_{\mathcal{X}}^{(h)}}_{(\mathbf{x}, \mathbf{n})} \cdot {\mathbf{A}_{\mathcal{Y}}^{(h)}}_{(\mathbf{y}, \mathbf{m})} \right) \mathbf{C}_{(\mathbf{n}, \mathbf{m})} {w}^{(h)} \\
    &= \sum_{h \in [{N_h}]} {\mathbf{A}}^{(h)}_{((\mathbf{x}, \mathbf{y}), :)} \mathbf{C} \ {w}^{(h)}.
\end{align}

Now consider the following Lemma:
\label{sec:lemma}
\smallbreak
\noindent \textbf{Lemma 1.} {\em Consider a bijective mapping of natural numbers, i.e., heads, onto 6-dimensional local displacements: $t(h): [{N_h}] \rightarrow \Delta(\mathbf{x}, \mathbf{y})$. Let ${\mathbf{A}}^{(h)} \in \mathbb{R}^{|\mathcal{X}||\mathcal{Y}| \times |\mathcal{X}||\mathcal{Y}|}$ be an attention matrix that holds the following:}
\begin{align}
    {\mathbf{A}}^{(h)}_{(\mathbf{x}, \mathbf{y}), (\mathbf{n}, \mathbf{m})} = \begin{cases} \mbox{1,} & \mbox{if $t(h) = (\mathbf{n}, \mathbf{m}) - (\mathbf{x}, \mathbf{y})$} \\ \mbox{0,} & \mbox{otherwise.} \end{cases} \label{eq:local_constraint_3s}
\end{align}
{\em Then, for any high-dimensional convolution with a kernel $K: \mathbb{R}^{6} \xrightarrow{} \mathbb{R}$, there exists $\{w^{(h)} \in \mathbb{R}\}_{h \in [N_h]}$ such that following equality holds:}
\begin{align}
    \text{Conv}(\mathbf{F}_{\mathcal{X}}, \mathbf{F}_{\mathcal{Y}})_{(\mathbf{x}, \mathbf{y})} = \sum_{h \in [{N_h}]} {\mathbf{A}}^{(h)}_{((\mathbf{x}, \mathbf{y}), :)} \mathbf{C} \ {w}^{(h)}.
\end{align}

{\em Proof.} \ Consider high-dimensional convolution at position $(\mathbf{x}, \mathbf{y})$:
\begin{align}
    \text{Conv}(\mathbf{F}_{\mathcal{X}}, \mathbf{F}_{\mathcal{Y}})_{(\mathbf{x}, \mathbf{y})} &\coloneqq \sum_{(\mathbf{n}, \mathbf{m}) \in \mathcal{N}(\mathbf{x}) \times \mathcal{N}(\mathbf{y})} \mathbf{C}_{(\mathbf{n}, \mathbf{m})} {K}([\mathbf{n} - \mathbf{x}, \mathbf{m}-\mathbf{y}]) \nonumber \\
    &= \sum_{(\pmb{\nu}, \pmb{\mu}) \in \Delta(\mathbf{x}, \mathbf{y})} \mathbf{C}_{(\mathbf{x}, \mathbf{y}) + (\pmb{\nu}, \pmb{\mu})} {K}((\pmb{\nu}, \pmb{\mu})) \nonumber \\
    &= \sum_{h \in [{N_h}]} \mathbf{C}_{(\mathbf{x}, \mathbf{y}) + t(h)} K(t(h)) \tag{ \ $t(h): [{N_h}] \rightarrow \Delta(\mathbf{x}, \mathbf{y})$ \ } \\
    &= \sum_{h \in [{N_h}]} \mathbf{C}_{(\mathbf{x}, \mathbf{y}) + t(h)} {w}^{(h)} \tag{ \ ${w}^{(h)} \coloneqq {K}(t(h)) \in \mathbb{R}$ \ } \\
    &= \sum_{h \in [{N_h}]} \left( \sum_{(\mathbf{n}, \mathbf{m}) \in \mathcal{X} \times \mathcal{Y}} \mathds{1}[t(h) = (\mathbf{n}, \mathbf{m}) - (\mathbf{x}, \mathbf{y})] \mathbf{C}_{(\mathbf{n}, \mathbf{m})} \right) {w}^{(h)} \nonumber \\
    &= \sum_{h \in [{N_h}]} {\mathbf{A}}^{(h)}_{((\mathbf{x}, \mathbf{y}), :)} \mathbf{C} {w}^{(h)}.
\end{align}

By applying Lemma 1, we conclude that the dot-product of Proxy Match Transform outputs is equivalent to the high-order convolution.  \QEDA

\section{Efficiency of Proxy Match Transform}
\label{sec:efficiency}

To demonstrate the superiority of the proposed PMT, we provide the efficiency comparison between different matchers, \eg, Geometric Transformer (GeoTr) by \citet{qin2022geotransformer} and Proxy Match Transform (PMT), both during training and inference phases in Tab.~\ref{table:efficiency}.
``Coarse-only'' and ``Coarse + Fine'' refer to two different Proxy Match TransformeR (PMTR) models with PMT integrated only at the coarse-level and both levels, respectively.
Specifically, we measure the computational efficiency by employing Floating Point Operations Per Second (FLOPS), and to assess the memory overhead and footprint, we record the peak memory usage for each method during both the training and inference phases, as well as the number of parameters.
We also provide the training/inference times required for each matcher.
For clarity in our comparison, when measuring the FLOPS, number of parameters, and train/inference times, we exclude those associated with the backbone and focus solely on the matchers: the coarse- or fine-level matcher.

\begin{table}[!h]
    \centering
    \caption{Efficiency comparison results between GeoTr~\cite{qin2022geotransformer} and PMT. Lower is better.}
    \label{table:efficiency}
    \resizebox{\textwidth}{!}{
	\begin{tabular}{l|cc|cccccc}
        \toprule
        \multirow{2}{*}{Method} & Coarse-level & Fine-level & FLOPS $\downarrow$ & \# Param. $\downarrow$ & Mem. train $\downarrow$ & Mem. test $\downarrow$ & Train time $\downarrow$ & Inference time $\downarrow$ \\
        & Matcher & Matcher & (G) & (K) & (GB) & (GB) & (ms) & (ms) \\
        \midrule
        GeoTransformer~\yrcite{qin2022geotransformer} & GeoTr & None & 9.67 & 926.85 & 6.96 & 3.10 & 8.93 & 8.04 \\
        PMTR (Coarse-only) & PMT & None &\textbf{0.45} & \textbf{273.85} & \textbf{2.12} & \textbf{0.28} & \textbf{4.06} & \textbf{3.23} \\
        \textbf{PMTR (Coarse + Fine)} & PMT & PMT & \underline{0.78} & \underline{296.15} & \underline{3.78} & \underline{0.88} & \underline{5.35} & \underline{3.75} \\
        
        \bottomrule
	\end{tabular}
    }
\end{table}

The results clearly indicate that PMT delivers substantial reductions not only in training/inference time but also in memory requirements.
Notably, PMT is approximately $\times \textbf{21.5}$ more efficient in FLOPS, needs $\times \textbf{3.4}$ more compact number of parameters, and $\times  \textbf{3.28}$ / $\times  \textbf{11.07}$ less required memory for training/inference phases compared to GeoTr.
Such efficiency is crucial, as it facilitates the practical deployment of our fine-level matcher for intricate matching tasks.

\newpage
\section{Additional implementation Details}
\label{sec:implementation}

\textbf{Attention Calculation.} 
We adopt the relative-position encoding strategy of PerViT~\citep{min2022peripheral} to compute the attention $\mathbf{A}_{\mathcal{X}}^{(h)}$.
Specifically, we compute pairwise Euclidean distances $\mathbf{R}_{\mathcal{X}} \in \mathbb{R}^{|\mathcal{X}| \times |\mathcal{X}|}$ each of which entry at position $\mathbf{q}, \mathbf{k} \in \mathbb{R}^{3}$ is defeind as $(\mathbf{R}_{\mathcal{X}})_{\mathbf{q},\mathbf{k}}=||\mathbf{q}-\mathbf{k}||_{2}$.
An MLP processes this to provide an attention score $\mathbf{A}_{\mathcal{X}}^{(h)}$.
$\mathbf{A}_{\mathcal{Y}}^{(h)}$ is similarly defined.
We refer the readers to the work of~\citet{min2022peripheral} for additional details.

\textbf{Model hyperparameters.} For the backbone network, we utilize KPConv-FPN~\citep{thomas2019kpconv} with subsampling radius of 0.01.
We leverage the global attention matrix for coarse-level matcher, and the local attention matrix (See Sec.~\ref{sec:impl}) for fine-level matchers.
The number of attention heads $N_h$ is set to 4.
Refer to Tab.~\ref{table:hyperparam} for the rest of the hyperparameters. Each matcher takes a specific input and output feature pair, applies a type of attention mechanism, and uses various hyperparameters crucial for its operation.

\vspace{-4mm}
\begin{table}[!ht]
    \centering
    \caption{Detail configurations and hyperparameters of different types of matchers. $\mathbf{A}_{\mathcal{Y}}$ similarly defined.}
    \label{table:hyperparam}
    \resizebox{\textwidth}{!}{
	\begin{tabular}{c|ccc|ccc}
        \toprule
        Matcher Type & Input Feature Pair & Output Feature Pair & Attention Type & $D_{\text{emb}}$ & $i$-th PMT & $D_{\text{proxy}}$ \\
        \midrule
        \multirow{2}{*}{Coarse-level matcher} & \multirow{2}{*}{$\{\mathbf{F}_{\mathcal{X}_1}, \mathbf{F}_{\mathcal{Y}_1}\}$} & \multirow{2}{*}{$\{\mathbf{F}_{\mathcal{X}_c}, \mathbf{F}_{\mathcal{Y}_c}\}$} & \multirow{2}{*}{global attention $\mathbf{A}_{\mathcal{X}_1}^{(h)}\in \mathbb{R}^{|\mathcal{X}_1|\times |\mathcal{X}_1|}$} & \multirow{2}{*}{512} & 1 & 32 \\
        &  &  &  &  & 2 & 128 \\
        \midrule
        \multirow{2}{*}{Fine-level matcher} & \multirow{2}{*}{$\{\mathbf{F}_{\mathcal{X}_2}, \mathbf{F}_{\mathcal{Y}_2}\}$} & \multirow{2}{*}{$\{\mathbf{F}_{\mathcal{X}_3}, \mathbf{F}_{\mathcal{Y}_3}\}$} & \multirow{2}{*}{local attention $\mathbf{A}_{\mathcal{X}_2}^{(h)}\in \mathbb{R}^{|\mathcal{X}_2|\times \epsilon}$} & \multirow{2}{*}{256} & 1 & 16 \\
        &  &  &  &  & 2 & 64 \\
        \midrule
        \multirow{2}{*}{Fine-level matcher} & \multirow{2}{*}{$\{\mathbf{F}_{\mathcal{X}_3}, \mathbf{F}_{\mathcal{Y}_3}\}$} & \multirow{2}{*}{$\{\mathbf{F}_{\mathcal{X}_f}, \mathbf{F}_{\mathcal{Y}_f}\}$} & \multirow{2}{*}{local attention $\mathbf{A}_{\mathcal{X}_3}^{(h)}\in \mathbb{R}^{|\mathcal{X}_3|\times \epsilon}$} & \multirow{2}{*}{128} & 1 & 8 \\
        &  &  &  &  & 2 & 32 \\
        \bottomrule
	\end{tabular}
    }
\end{table}

\section{Evaluation Metrics}
\label{sec:metric}

We employ four different metrics to assess the results. Consider a pair of input point clouds $\{\mathcal{X},\mathcal{Y}\}$.
The ground truth SE(3) relative pose between the point clouds is represented by $\{\mathbf{R}^{\mathrm{GT}},\mathbf{t}^{\mathrm{GT}}\}$,
while the prediction is denoted as $\{ \mathbf{R}, \mathbf{t}\}$. We define $\mathbf{T}(\cdot)$ as a function that transforms input pose with corresponding rotation $\mathbf{R}$ and translation $\mathbf{t}$.

\textit{Chamfer Distance (CD).} The chamfer distance between two point clouds $S_1,S_2$ is defined as
\begin{equation}
    d_{\text{CD}}(S_1,S_2)=\frac{1}{S_1}\sum_{x\in S_1}\min_{y\in S_2}\|x-y\|_2^2 + \frac{1}{S_2}\sum_{y \in S_2}\min_{x\in S_1}\|x-y\|_2^2,
\end{equation}
which measures the sum of the distance between nearest neighbor correspondences between point clouds. To assess the quality of shape assembly, we measure the chamfer distance between ground truth assembly and the prediction as:
\begin{equation}
    \text{CD}=d_{\text{CD}}(\mathbf{T}(\mathcal{X})\cup\mathcal{Y},\mathbf{T}^{\mathrm{GT}}(\mathcal{X})\cup\mathcal{Y}).
    \label{eq:cd}
\end{equation}

\textit{CoRrespondence Distance (CRD)}. 
While the Chamfer distance calculates the distance between two point clouds, its ability to capture more complex features of the object's geometry, such as symmetry and rotation, is limited. To overcome this limitation, we define a new metric, CoRrespondence Distance (CRD). CRD is simply defined as the Frobenius norm between two point clouds:

\begin{equation}
    \text{CRD}=\frac{1}{L}\sum_{i=1}^{L} \|(\mathbf{T}(\mathcal{X})\cup\mathcal{Y})_{i} - (\mathbf{T}^{\mathrm{GT}}(\mathcal{X})\cup\mathcal{Y})_{i}\|_F, \\
    \label{eq:crd}
\end{equation}
where $L=|\mathcal{X}|+|\mathcal{Y}|$ is the size of assembled object.
By considering all pairwise distances between point clouds, it offers a more comprehensive measure of similarity, capturing both proximity and structural alignment:

\textit{Rotational-, Translational-RMSE (RMSE(R), RMSE(T))}. Finally, to directly measure the prediction accuracy of transformation parameters, we compute the root mean square error (RMSE) between predicted and ground-truth rotation and translation, respectively. Following the protocols of \citet{sellan2022breakingbad}, we use Euler angle representation for rotation:

\begin{equation}
    \text{RMSE(R)} = \frac{1}{\sqrt{3}} \|\mathbf{R}-\mathbf{R}^{\mathrm{GT}}\|_{F}, \ \ \ \ \ \
    \text{RMSE(T)} = \frac{1}{\sqrt{3}} \|\mathbf{t}-\mathbf{t}^{\mathrm{GT}}\|_{F}.
    \label{eq:rmse}
\end{equation}

\textbf{Additional metrics for multi-part assembly.} Also, we employ two additional metrics to evaluate multi-part assembly performance. Consider a set of input point clouds $\mathcal{P}=\{\mathcal{P}_i \}_{i=1}^{P}$ with $P$ parts. The ground truth SE(3) relative poses between the point clouds are represented by $\{\mathbf{T}^{\mathrm{GT}}_i\}_{i=1}^{P}$,
while the prediction is denoted as $\{\mathbf{T}_i\}_{i=1}^{P}$.
Similar to pairwise assembly, the assembled object can be represented with $\bigcup_{i=1}^{P}\mathbf{T}_i(\mathcal{P}_i)$.
Note that in our context, the direction of the pose is defined as the transformation that aligns each part $\mathcal{P}_{i}$ with the coordinate frame of the largest fracture as an anchor.

\textit{Part Accuracy (Chamfer Distance-based)}. Part accuracy (PA)~\citep{li2020learning} is defined as the percentage of fractures with Chamfer Distance (CD) less than the predefined threshold $\tau_{\text{CD}} = 0.01$: 
\begin{equation}
    \text{PA}_{\text{CD}}= \frac{1}{P}\sum_{i=1}^{P}\mathbbm{1}\left( d_{\text{CD}}(\mathbf{T}_i(\mathcal{P}_{i}),\mathbf{T}^{\mathrm{GT}}_{i}(\mathcal{P}_{i}))< \tau_{\text{CD}} \right).
    \label{eq:pa_cd}
\end{equation}
\textit{Part Accuracy (Correspondence Distance-based)}. Our proposed CoRrespondence Distance (CRD) can be seamlessly adapted for Part Accuracy (PA) evaluation. This adaptation involves substituting the Chamfer Distance (CD)  with the CoRrespondence Distance (CRD), and setting the threshold $\tau_{\text{CRD}} = 0.1$:
\begin{equation}
    \text{PA}_{\text{CRD}}= \frac{1}{P}\sum_{i=1}^{P}\mathbbm{1}\left( \frac{1}{|\mathcal{P}_i|} \sum_{j=1}^{|\mathcal{P}_i|} \|{\mathbf{T}_i(\mathcal{P}_{i})}_{j}-{\mathbf{T}^{\mathrm{GT}}_{i}(\mathcal{P}_{i})}_{j}\|_{F}< \tau_{\text{CRD}} \right).
    \label{eq:pa_crd}
\end{equation}

\newpage
\section{Additional Qualitative Results}

\begin{figure*}[ht]
\vskip 0.1in
\begin{center}
\includegraphics[width=\textwidth]{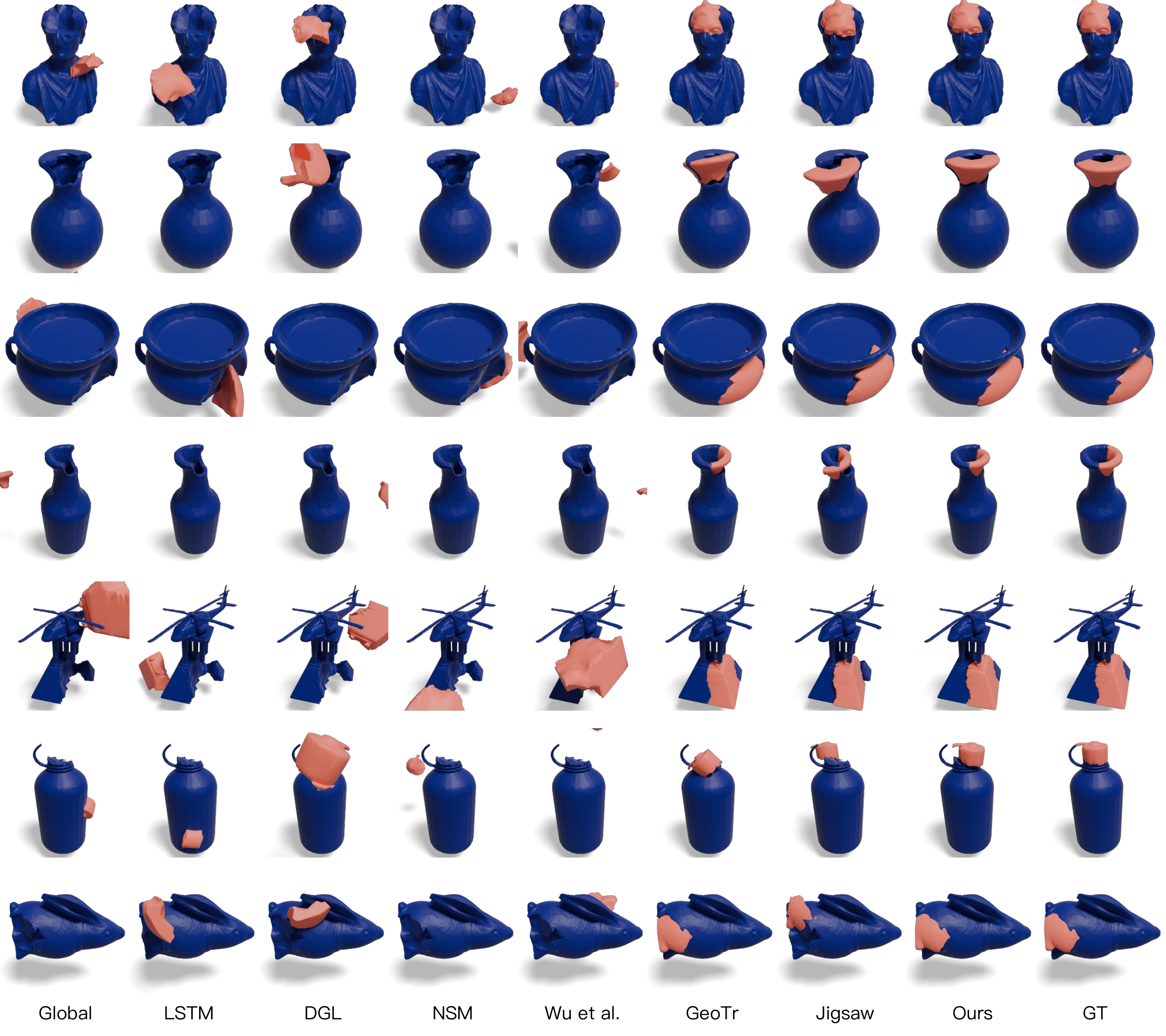}
\vspace{-5mm}
\caption{Additional qualitative results of pairwise shape assembly on Breaking Bad dataset.}
\label{fig:qual_supp_pairwise}
\end{center}
\vskip -0.2in
\end{figure*}

\begin{figure*}[ht]
\vskip 0.1in
\begin{center}
\includegraphics[width=\textwidth]{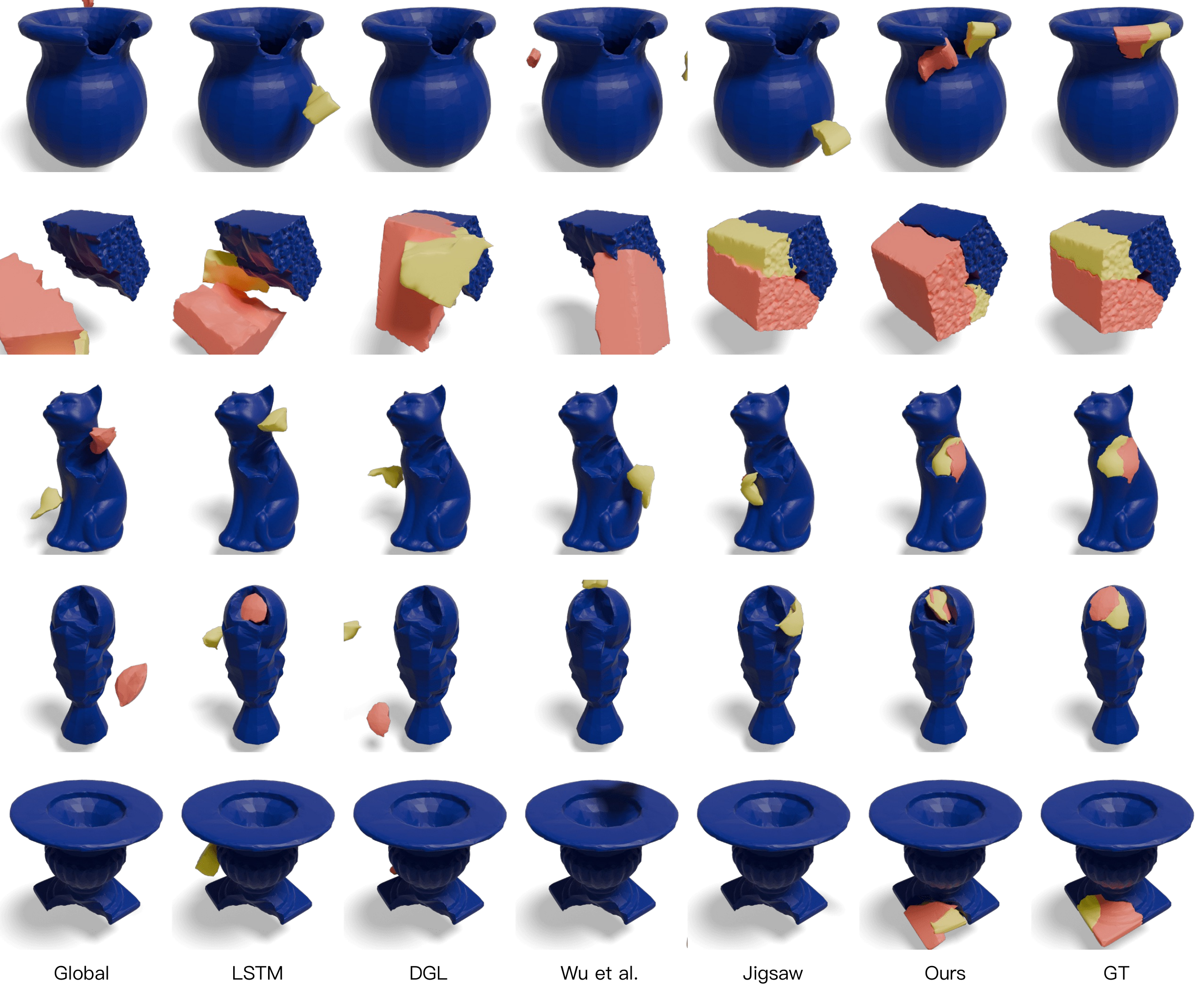}
\vspace{-5mm}
\caption{Additional qualitative results of multipart shape assembly on Breaking Bad dataset.}
\label{fig:qual_supp_mpa}
\end{center}
\vskip -0.2in
\end{figure*}

\end{document}